\documentclass{article}

\usepackage[preprint]{neurips_2026}
\usepackage{multirow}

\usepackage[utf8]{inputenc} % allow utf-8 input
\usepackage[T1]{fontenc}    % use 8-bit T1 fonts
\usepackage{hyperref}       % hyperlinks
\usepackage{url}            % simple URL typesetting
\usepackage{booktabs}       % professional-quality tables
\usepackage{amsfonts}       % blackboard math symbols
\usepackage{amsmath}
\usepackage{nicefrac}       % compact symbols for 1/2, etc.
\usepackage{microtype}      % microtypography
\usepackage{xcolor}         % colors
\usepackage{graphicx}

\title{RoPEMover: Depth-Aware Object Relocation via Positional Embeddings}

\author{%
  Ipek Oztas \\
  Bilkent University, Brown University \\
  \texttt{ipek\_oztas@brown.edu} \\
  \And
  Duygu Ceylan \\
  Adobe Research \\
  \texttt{ceylan@adobe.com} \\
  \And
  Aybars Bugra Aksoy \\
  Bilkent University \\
  \texttt{bugra.aksoy@ug.bilkent.edu.tr} \\
  \And
  Aysegul Dundar \\
  Bilkent University \\
  \texttt{adundar@bilkent.edu.tr} \\
}

\begin{document}
\maketitle
% \begin{figure}[h]
%     \centering
%     \fbox{
%         \begin{minipage}[c][8cm][c]{0.95\linewidth}
%             \centering
%             \vspace{2cm}
%             \textbf{Teaser Figure Placeholder}
%             \vspace{2cm}
%         \end{minipage}
%     }
%     \caption{Our method enables realistic object displacement in a single image while correctly handling occlusions, revealing unseen regions, and preserving shadows and reflections. By extending RoPE to a depth-aware formulation, we explicitly control object ordering, allowing placement in front of or behind other scene elements. Beyond motion, the same framework supports downstream edits such as object removal and adding objects to a scene, achieving consistent illumination, shadow casting, and overall scene integration.}
%     \label{fig:teaser}
% \end{figure}

\begin{figure}[h]
    \centering
    \includegraphics[width=1.0\linewidth, trim=3.5cm 3cm 11cm 0cm,
                     clip]{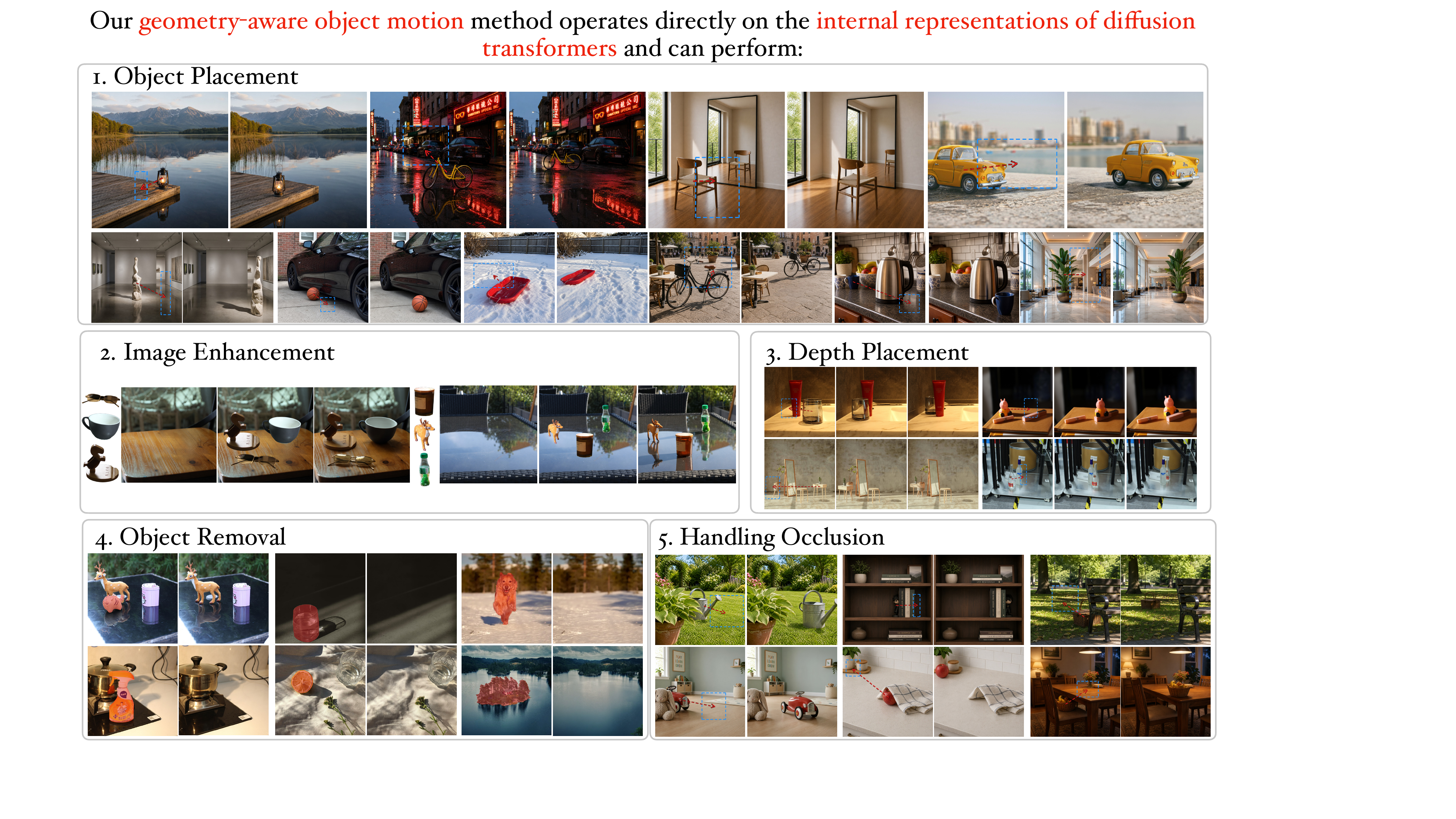}
    \caption{Our method enables realistic object displacement in a single image while correctly handling occlusions, revealing unseen regions, and preserving shadows and reflections. By extending RoPE to a depth-aware formulation, we explicitly control object ordering, allowing placement in front of or behind other scene elements. Beyond motion, the same framework supports downstream edits such as object removal and adding objects to a scene, achieving consistent illumination, shadow casting, and overall scene integration.}
    \label{fig:teaser}
\end{figure}

\begin{abstract}
Moving an object in a single image requires geometry-consistent spatial rearrangement, including handling occlusions, revealing previously unseen regions, and maintaining coherent shadows and reflections. Existing approaches are not well suited to this setting and often fail to preserve such scene-level consistency.
We address this problem by introducing a geometry-aware object motion method that operates directly on the positional representations of diffusion transformers. Our key insight is that rotary positional embeddings (RoPE) define a structured spatial field that can be explicitly manipulated to induce controlled motion. We extend 2D RoPE into a depth-aware formulation that encodes 3D spatial structure, enabling consistent object displacement and scene-aware updates.
Our model is trained using synthetic data combined with a small set of real images via parameter-efficient fine-tuning. Despite minimal real supervision, it preserves object identity under large spatial displacements, generates plausible content in newly revealed regions, and consistently updates scene-dependent effects such as shadows and illumination.
Experimental results on standard object motion benchmarks demonstrate state-of-the-art performance across all evaluation metrics. \footnote{Project page: \url{https://ipekoztas.github.io/RoPEMover/}}
\end{abstract}
\section{Introduction}

Moving an object within a single image is a common yet challenging task in image editing, with applications in content creation, visual effects, and interactive design. Given an input image and a target location, the goal is to reposition an object while preserving its appearance and maintaining consistency with the surrounding scene. This requires more than spatial displacement: the object must adapt to perspective and scale, previously occluded regions must be plausibly completed, and interactions with the scene must remain coherent. In particular, object motion can affect scene-dependent elements such as shadows and illumination, which must update consistently with the new configuration. These requirements make object motion a geometry-aware transformation problem rather than a purely appearance-based edit.

A straightforward strategy for object movement is to decompose the task into object removal at the source location and insertion at the target location \cite{yang2023paint, chen2024anydoor, chen2024zero, yu2025omnipaint}. However, such two-step approaches often fail to preserve consistency: accurately removing an object requires accounting for associated effects such as shadows and reflections, while insertion models frequently alter object identity or produce mismatched lighting and geometry. Similarly, copy-and-paste based methods followed by harmonization struggle to account for perspective changes and typically fail to produce coherent scene interactions \cite{winter2024objectdrop, alzayer2025magic, zhu2025training}.
To address these challenges, recent generative approaches rely on large-scale training data, often constructing dedicated synthetic datasets and data generation pipelines to model object movement \cite{ruan2026ctrl}.  A separate line of work leverages video diffusion models \cite{yu2025objectmover} to capture motion priors, but this introduces additional complexity and requires substantial training data. Despite these advances, achieving precise, geometry-aware object motion within a single image remains challenging.

In this work, we take a different approach and enable object motion by directly manipulating the internal positional representations of diffusion transformers, rather than relying on large-scale motion data or temporal priors. Our key insight is that rotary positional embeddings (RoPE) encode a structured spatial field that can be transformed to induce controlled object displacement. By warping these embeddings according to target motion and augmenting them with depth-guided conditioning, our method enables geometry-aware object movement within a single image. Leveraging the strong spatial priors of pretrained models, this formulation allows us to learn effective motion behavior from simple synthetic data combined with a small set of real images. Despite this minimal supervision, the model achieves precise spatial control, maintains consistency in both visible and newly revealed regions, and naturally propagates motion to scene-dependent effects such as shadows and illumination.

We evaluate our method on standard image object motion benchmarks and compare with a variety of baselines. We summarize our contributions as follows.

\begin{itemize}
    \item \textbf{Geometry-aware object motion via representation manipulation.}
    We introduce a method that enables object movement by directly manipulating rotary positional embeddings (RoPE) within diffusion transformers, providing explicit spatial control without requiring explicit 3D reconstruction or video-based modeling.

   \item \textbf{Depth-guided spatial transformation for consistent edits.}
We augment RoPE-based manipulation with depth-aware conditioning, enabling geometry-consistent object displacement that preserves object identity, handles occlusions, and generates coherent content in newly revealed regions. This formulation allows controlled manipulation of depth ordering enabling objects to move both in front of and behind other scene elements. This is a capability that prior methods often fail to achieve.

    \item \textbf{Data-efficient learning of motion behavior.}
    By leveraging the strong spatial priors of pretrained diffusion models, our approach learns effective object motion from simple synthetic data and a small set of real images, avoiding the need for large-scale motion datasets or complex data generation pipelines.

    \item \textbf{State-of-the-art performance with physically consistent effects.}
    Our method achieves superior results in object motion tasks, producing high-fidelity edits with consistent propagation of scene-dependent effects such as shadows and illumination, outperforming prior approaches in both visual quality and geometric consistency.
\end{itemize}
\section{Related Work}

\paragraph{Image Editing for Object Relocation.}
A common approach for moving objects in images decomposes the task into object relocation followed by image completion. Early methods based on reference-guided insertion~\cite{yang2023paint, chen2024anydoor, yu2025omnipaint} paste an object from a source image into a new location, but are not designed for object relocation and often fail to preserve object identity. This issue becomes particularly pronounced for generative motion, where even small discrepancies are perceptually salient.
More recent methods explicitly target object relocation. Approaches such as FreeFine~\cite{zhu2025training} and related drag-based methods~\cite{winter2024objectdrop, alzayer2025magic, lu2025inpaint4drag} perform geometric transformation followed by inpainting and refinement. While effective for moderate edits, these pipelines frequently exhibit a copy-paste effect, failing to model scene-dependent interactions such as shadows, reflections, and indirect illumination. As a result, edited objects often appear physically inconsistent, e.g., floating or poorly grounded.
ObjectMover~\cite{yu2025objectmover} and ChronoEdit \cite{wu2026chronoedit} instead leverage video diffusion models to learn motion-aware priors, but at the cost of expensive data construction and training. Despite this, such approaches can still exhibit inconsistencies, including duplicated or missing objects.
More recently, instruction-tuned diffusion transformers such as Flux Kontext~\cite{labs2025flux1kontextflowmatching} and QwenEdit~\cite{wu2025qwenimagetechnicalreport} unify generation and editing within a single diffusion backbone. While these models enable editing via natural language, control over spatial layout remains imprecise: prompts specifying object motion are interpreted loosely.

In contrast, video-based approaches rely on heavy temporal modeling, while instruction-based methods provide only coarse spatial control. Our method instead builds directly on the internal representations of instruction-tuned diffusion transformers, reusing their learned spatial structure (including rotary positional embeddings) to enable precise geometric manipulation. This allows explicit and controllable object motion within a single image, without video supervision or large-scale motion datasets, while maintaining strong physical consistency and coherent scene interactions.

\paragraph{Applications built on RoPE embedding.}
RoPE has become a standard positional encoding in large language, vision, and diffusion transformer models due to its ability to encode relative positions in attention \cite{su2024roformer, weivideorope, labs2025flux1kontextflowmatching, wu2025qwenimagetechnicalreport}.
Recent work has shown that pretrained models using RoPE can be repurposed for controllable generation by directly manipulating their positional representations. For instance, DitFlow~\cite{pondaven2025video} and RoPECraft~\cite{gokmenropecraft2025} leverage RoPE in pretrained video diffusion transformers to enable motion transfer via trajectory-guided optimization. More recently, RoPE-Infinity~\cite{yesiltepe2025infinity} demonstrates that reparameterizing temporal RoPE enables controllable long-horizon and effectively infinite video generation at inference time.
In contrast, we explore RoPE manipulation for geometry-aware object relocation in single images. The concurrent PE-Field~\cite{bai2025positional} work extends 2D positional encodings to a 3D depth-aware field primarily targeting novel view synthesis. In contrary, our method repurposes the unused temporal RoPE axis within an instruction-tuned model and targets object relocation. Our method explicitly models occlusion ordering, shadow propagation, and scene-consistent inpainting.
\section{Method}\label{sec:method}

\subsection{Preliminaries}

\textbf{Rotary Positional Encoding (RoPE)}
~\cite{su2024roformer} encodes positional information by applying a position-dependent rotation to query and key features, enabling self-attention to model relative spatial relationships. Given a token at position $m$ with feature vector $\mathbf{x}_m \in \mathbb{R}^d$, RoPE splits the vector into $d/2$ pairs, each interpreted as a complex number $\mathbf{z}^{(i)}_m = \mathbf{x}^{(2i-1)}_m + j\,\mathbf{x}^{(2i)}_m$. A rotation $\Phi_{m,i} = e^{j m \theta^{-2i/d}}$ is then applied, where $\theta$ is a base frequency. This embeds positional information into phase while preserving relative positional structure through inner products of rotated queries and keys.
In diffusion transformers, this mechanism induces a structured spatial representation within the attention layers. In our work, we leverage this property as a manipulable spatial field for geometry-aware object relocation.

\begin{figure}
    \centering
    \includegraphics[width=1.0\linewidth,
                     trim=0cm 0cm 0cm 0cm,
                     clip]{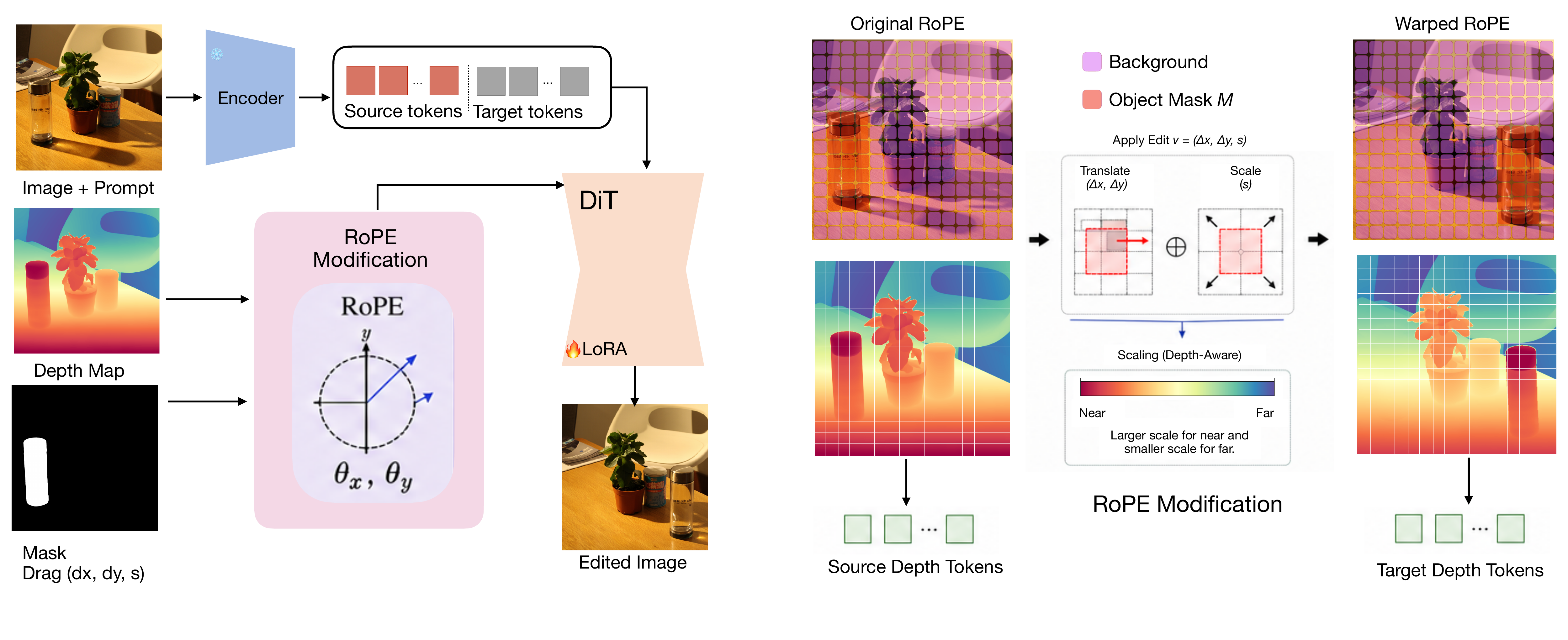}
    \caption{Overview of the method: Given an input image, object mask, and a user-specified drag signal $(d_x, d_y, s)$, we encode the desired motion by warping the Rotary Positional Embeddings (RoPE) of tokens within the object region. A depth map is estimated and used to construct a depth-aware 3D RoPE representation, where depth is injected along the temporal axis. The modified tokens are processed by a diffusion transformer (DiT) to generate the edited image, enabling geometry-consistent object relocation with correct scaling and occlusion handling.}
    \label{fig:method}
\end{figure}

\subsection{Overview}
Given a single input image, an object mask $\mathbf{M_\text{src}}$ identifying the region to be moved, and a user-specified drag signal $(d_x,d_y,s)$ encoding the desired displacement and scale change, our method produces an edited image in which the object has been relocated in a geometry-consistent manner. Our method first warps the RoPE encodings of masked tokens according to the drag signal, directly inducing the desired spatial transformation within the attention layers of the diffusion transformer (Section~\ref{sec:drag}). Second, we estimate a metric depth map from the input image and modify it based on the desired object displacement and inject it into the unused temporal axis of the factorized 3D RoPE, providing the model with explicit geometric context about the scene (Section~\ref{sec:depth}). Together, these modifications guide the model to synthesize the object at the target location with correct occlusion handling, plausible completion of newly revealed regions, and consistent updates to scene-dependent effects such as shadows and reflections.
 
\subsection{Drag Signal Encoding}
\label{sec:drag}
We encode the user-specified drag as a triplet $(d_x, d_y, s)$, where $(d_x, d_y)$ are pixel-space displacement vectors from the object's source position to its desired target position, and $s$ is a scale ratio indicating the relative size change of the object.

% \paragraph{Spatial offsets.}
% The drag displacement $(d_x, d_y)$ is defined in the coordinate frame of the original image and is rescaled proportionally when the image is resized to the working resolution $W \times H$:
% \begin{equation}
%     \tilde{d}_x = d_x \cdot \frac{W}{W_0}, \quad \tilde{d}_y = d_y \cdot \frac{H}{H_0},
% \end{equation}
% where $W_0 \times H_0$ is the original image resolution. Since the diffusion transformer operates on non-overlapping $16 \times 16$ patches, the pixel offsets are converted to token-grid offsets:
% \begin{equation}
%     \delta_x = \left\lfloor \frac{\tilde{d}_x}{p} \right\rceil, \quad
%     \delta_y = \left\lfloor \frac{\tilde{d}_y}{p} \right\rceil, \quad p = 16.
% \end{equation}

\paragraph{Spatial offsets.}
The drag displacement $(d_x, d_y)$, defined in the coordinates of the original $W_0 \times H_0$ image, is rescaled to the working resolution $W \times H$ and converted to token-grid offsets on the $p \times p$ patch grid ($p = 16$):
\begin{equation}
    \delta_x = \left\lfloor \frac{d_x}{p} \cdot \frac{W}{W_0} \right\rceil, \quad
    \delta_y = \left\lfloor \frac{d_y}{p} \cdot \frac{H}{H_0} \right\rceil.
\end{equation}

\paragraph{Scale ratio.}
To capture whether the object should appear larger or smaller at the target location, we compute a scale ratio from the binary object masks during training. Let $A_\text{src}$ and $A_\text{tgt}$ denote the foreground pixel counts of the source mask and target mask, both measured at a common resolution of $512 \times 512$ to ensure comparability across datasets. The scale ratio is defined as:
\begin{equation}
    s = \sqrt{\frac{A_\text{tgt}}{A_\text{src}}},
\end{equation}
which corresponds to the linear scale factor under the assumption of uniform area scaling.
% We discretise $s$ into three categories with threshold $\tau = 0.15$: \emph{smaller} ($s < 1 - \tau$), \emph{same} ($|s - 1| \leq \tau$, clamped to $s = 1$), and \emph{larger} ($s > 1 + \tau$).

\paragraph{RoPE warp.}
The drag signal is applied by warping the Rotary Position Embeddings (RoPE) of the image tokens that lie within the source object mask. Let $(r, c)$ denote the row and column indices of a token on the $H_t \times W_t$ token grid, and let $(c_h, c_w)$ be the centroid of the mask in token coordinates. For each masked token, we replace its RoPE encoding with that of a position computed based on the intended transformation: 
\begin{equation}
    r' = (r - c_h) \cdot s + c_h + \delta_y, \quad
    c' = (c - c_w) \cdot s + c_w + \delta_x,
\end{equation}
clamped to the valid grid range $[0, H_t{-}1] \times [0, W_t{-}1]$. Intuitively, this causes the masked tokens to ``behave as if'' they were located at the dragged, scaled position, guiding the model to synthesize the object at the target location as shown in Fig. \ref{fig:method}. The warp is applied only during the first $t=20$ denoising steps, as structural layout is determined early in the reverse diffusion process (see Fig. \ref{fig:timestep}).

% TODO: figure showing the RoPE warp on the token grid

\subsection{Depth-Aware 3D RoPE}
\label{sec:depth}
\paragraph{Motivation.}
The factorized 3D RoPE variant used in Qwen-Image \cite{wu2025qwenimagetechnicalreport} assigns each image (latent) token a three-dimensional position index $(t, h, w)$, where $t$ encodes a virtual temporal/context position. For single-image inputs, all target tokens share $t = 0$, with non-zero $t$ values reserved as constant offsets that separate context image tokens from target tokens while preserving their internal spatial structure. For single-image editing, this temporal axis is otherwise unused on the target side. We repurpose this axis as a \emph{depth} channel to inject metric depth information, providing the model with explicit 3D geometric context about the scene.

\paragraph{Depth estimation and normalization.}
For training, we estimate metric depth for both source and target images using MoGe-2~\cite{wang2025moge2accuratemonoculargeometry}, a monocular geometry estimator that produces dense depth maps. Because MoGe outputs are only determined up to an unknown per-image affine transformation, we perform scene-level affine alignment before computing depth offsets. Specifically, we fit
\begin{equation}
D_\text{tgt} \approx \alpha D_\text{src} + \beta
\end{equation}
via least-squares regression on \emph{background} pixels (those belonging to neither the source nor target object mask). The aligned target depth is then
\begin{equation}
\tilde{D}\text{tgt} = \frac{D\text{tgt} - \beta}{\alpha},
\end{equation}
which brings the target depth into the same metric scale as the source. All raw depth maps are normalized into the range $[1, 2]$, avoiding sign ambiguities in the complex RoPE exponentials.
% \begin{equation}
% \hat{D} = \frac{D - D_{\min}}{D_{\max} - D_{\min} + \epsilon} + 1, \quad \epsilon = 10^{-8}.
% \end{equation}
% The offset $+1$ ensures all values are strictly positive, avoiding sign ambiguities in the complex RoPE exponentials.

\paragraph{Injection into RoPE.}
The Qwen-Image DiT uses factorized 3D RoPE with axis dimensions $[16, 56, 56]$ for the temporal, height, and width axes, respectively. The position encoding function computes complex exponentials $e^{i \theta_k p}$ where $p$ is the position index and $\theta_k$ are frequency bases, yielding 8, 28, and 28 complex frequency components per axis (64 total per token). %These are concatenated and applied as element-wise rotations to the query and key vectors in every attention layer.

The normalized depth map $\hat{D} \in [1, 2]$ is downsampled to the token grid resolution $H_t \times W_t = H/16 \times W/16$ via bilinear interpolation. The per-token depth value $\hat{D}{r,c}$ is then written into the first complex component of the temporal axis of the RoPE frequency tensor:
\begin{equation}
\mathbf{f}\text{img}[\ell, 0] \leftarrow \hat{D}{r,c},
\end{equation}
where $\ell = r \cdot W_t + c$ is the linearized token index and $\mathbf{f}\text{img} \in \mathbb{C}^{L \times 64}$ is the image RoPE frequency tensor. 
%This injection is performed for the target image tokens (segment0) and the source image tokens (segment1). We explore two configurations:

\paragraph{Inference time target depth synthesis.}
At inference, only the source image is available. We construct a target depth map from the source depth $D_\text{src}$ using the drag parameters $(dx, dy)$ and a user provided per-object depth offset $\Delta z$. First, we inpaint the depth of the object region by solving the Laplace equation $\nabla^2 D = 0$ with Dirichlet boundary conditions at the mask boundary. We apply a safety margin of 3 pixels to dilate the mask. We replace the contaminated border pixels with their nearest clean background values via Euclidean distance transform before solving the Laplace equantion, to avoid depth bleeding from the object edges. This results in $D_\text{bg}$, a depth map with a smooth background surface. We then transform the source mask by the drag triplet $(dx, dy, s)$ to obtain $M_\text{tgt}$. Specifically, for each pixel $(i, j) \in M_\text{tgt}$, the corresponding source pixel is found at $(i - dy, j - dx)$. If this falls inside $M_\text{src}$, the object depth is copied with the depth offset applied:
\begin{equation}
D_\text{obj}(i, j) = D_\text{src}(i - dy, j - dx) + \Delta z.
\end{equation}
Pixels that do not map back to a valid source location fall back to $\operatorname{median}(D_\text{src}[M_\text{src}]) + \Delta z$. This per-pixel warp preserves the object's internal depth variation (e.g.\ the curvature of a chair seat), unlike the flat-offset approach.
Next, we composite the depth pf the moved object onto the background using a z-buffer depth test:
\begin{equation}
D_\text{out}(i,j) = \begin{cases}
D_\text{obj}(i,j) & \text{if } D_\text{obj}(i,j) < D_\text{bg}(i,j), \\
D_\text{bg}(i,j),  & \text{otherwise}
\end{cases}
\quad (i,j) \in M_\text{tgt},
\end{equation}
where smaller depth values correspond to surfaces closer to the camera. This ensures correct occlusion handling when the object moves behind existing scene geometry.
Finally, we apply a distance-weighted blend in a narrow band around the target mask boundary where the object is in front by smoothing the depth discontinuity:
\begin{equation}
D_\text{blend}(i,j) = w(i,j) \cdot D_\text{out}(i,j) + \bigl(1 - w(i,j)\bigr) \cdot D_\text{bg}(i,j), \quad w(i,j) = \operatorname{clamp}\!\left(\frac{d_\text{EDT}(i,j)}{r}, 0, 1\right).
\end{equation}
We apply the blending for $(i,j) \in \mathcal{B}$, where $\mathcal{B}$ is the set of pixels within the dilated-minus-eroded boundary band of $M_\text{tgt}$ for which the object is in front ($D_\text{out} < D_\text{bg}$), $d_\text{EDT}$ is the Euclidean distance transform of $M_\text{tgt}$, and $r = 3\,\text{px}$ is the blend radius. Pixels outside $\mathcal{B}$ retain their value from $D_\text{out}$. The synthesized depth is then normalized to the range $[1, 2]$ identically to the source depth. (See Fig. \ref{fig:depth_synthesis})

\subsection{Data Generation and Training}
We train our model using a combination of simple synthetic data and a small set of real images. 
For synthetic data, we build on the CLEVR dataset~\cite{johnson2016clevrdiagnosticdatasetcompositional}, generating paired scenes composed of basic geometric primitives in which a single object is translated while all other factors remain fixed. We curated three datasets of 2{,}000, 5{,}000, and 10{,}000 paired scenes from CLEVR.
This produces clean supervision for object motion and disocclusion without requiring realistic rendering or large-scale curated datasets.
To complement this, we collect a small real-image dataset consisting of 165 paired photographs in which an object is manually repositioned within the same scene under consistent viewpoint and lighting. 
This provides real-world examples of object motion and disocclusion, bridging the appearance gap between synthetic and real images. %\ceylan{Add quantity of each dataset}

\begin{figure}
    \centering
    \includegraphics[width=1.0\linewidth]{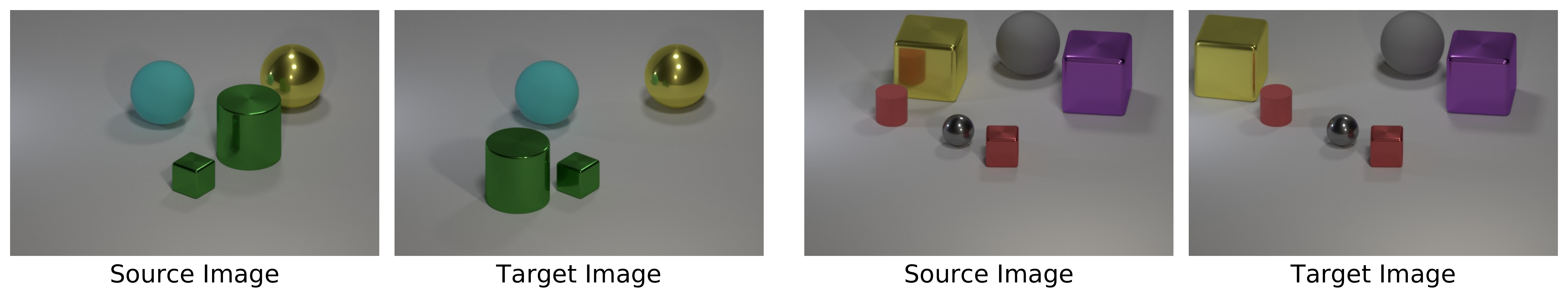}
    \caption{An example source-target pair from our synthetic CLEVR split. One object moves, with all other geometry, lighting, and materials held fixed. }
    \label{fig:clevr_example}
\end{figure}

We train LoRA adapters in two stages. We first train on the synthetic data to learn precise and controllable edit behavior. 
We then perform a lightweight finetuning stage on the real-image dataset to improve robustness and visual fidelity.
Despite the simplicity of the synthetic data as shown in Fig. \ref{fig:clevr_example}, our method generalizes well to real-world images. 
We attribute this to the strong internal representations of the underlying diffusion transformer, which already encode rich visual and geometric priors. 
In this context, synthetic data serves primarily to localize and parameterize edits, rather than modeling from scratch.
Full details of data generation, real data collection, and training setup are provided in the supplementary material.

\section{Results}\label{sec:results}

\subsection{Metrics and Dataset}

For evaluation, we use two datasets introduced by \cite{yu2025objectmover}: ObjMove-A and ObjMove-B. ObjMove-A consists of 200 captured image sets, where each set provides two views of the same scene differing only in the position of a single object, together with a clean background image of the scene with the object absent. Object masks are obtained via SAM \cite{sam}. Because each set supplies a ground-truth source–target pair, ObjMove-A enables quantitative comparison between the generated image and the reference target, and we use it for all numerical evaluations. In contrast, ObjMove-B does not provide paired ground truth and is thus used for the user study presented in the supplementary material.
Following \cite{yu2025objectmover,  tarres2024thinkingoutsidebboxunconstrained}, we adopt object-level metrics — DINO-Score \cite{caron2021emergingpropertiesselfsupervisedvision}, CLIPScore \cite{radford2021learningtransferablevisualmodels}, and DreamSim \cite{fu2023dreamsimlearningnewdimensions} — computed on both the cropped target and source regions. The target crop measures how well the moved object is preserved at its new location, while the source crop measures how completely it is removed from the original one. We additionally evaluate the metrics on the background region (the complement of the target region) to assess how faithfully the rest of the scene is preserved. Finally, we report PSNR over the full generated image as a measure of overall image similarity.

%en iyi ikinci metod underline
\begin{table*}[t]
\centering
\caption{Comparison on the ObjMove-A benchmark. 
For CLIP, DINO, and PSNR scores, higher-is-better; for DreamSim lower-is-better.}
\label{tab:comparison_test_split}
\resizebox{\textwidth}{!}{%
\begin{tabular}{l ccc ccc ccc c}
\toprule
 & \multicolumn{3}{c}{\textbf{CLIP Score} $\uparrow$} 
 & \multicolumn{3}{c}{\textbf{DINO Score} $\uparrow$}
 & \multicolumn{3}{c}{\textbf{DreamSim} $\downarrow$}
 & \textbf{PSNR} $\uparrow$ \\
\cmidrule(lr){2-4}\cmidrule(lr){5-7}\cmidrule(lr){8-10}
\textbf{Method} & Tgt & BG & Src & Tgt & BG & Src & Tgt & BG & Src & (dB) \\
\midrule
ChronoEdit   \cite{wu2026chronoedit}    & 73.27 & 92.93 & 77.65 & 45.27 & 91.17 & 44.28 & 0.6261 & 0.1026 & 0.5896 & 20.49 \\
3DiT       \cite{michel2023object}      & 75.74 & 91.08 & 83.24 & 48.78 & 86.81 & 48.21 & 0.5723 & 0.1711 & 0.4934 & 18.59 \\
DragAnything  \cite{wu2024draganything}   & 76.51 & 82.26 & 84.28 & 48.85 & 66.69 & 36.21 & 0.4460 & 0.3681 & 0.5233 & 11.21 \\
DragDiffusion  \cite{shi2024dragdiffusion}  & 74.40 & 91.35 & 80.00 & 46.51 & 86.97 & 46.93 & 0.6237 & 0.1369 & 0.5741 & 19.56 \\
Inpaint4Drag  \cite{lu2025inpaint4drag}   & {92.08} & 96.49 & 83.46 & {85.09} & 95.10 & 55.54 & {0.1379} & 0.0632 & 0.4798 & 22.85 \\
MagicFixup  \cite{alzayer2025magic}     & 89.62 & 96.06 & 92.60 & 80.90 & 95.29 & {78.49} & 0.1755 & 0.0552 & {0.1853} & 23.32 \\
GeoDiffuser   \cite{sajnani2025geodiffuser}   & 76.13 & 93.76 & 86.64 & 53.18 & 91.88 & 59.73 & 0.5430 & 0.0987 & 0.4109 & 19.65 \\
Qwen Image \cite{wu2025qwenimagetechnicalreport} & 81.23 & 94.01 & 81.23 & 64.14 & 92.12 & 56.78 & 0.4132 & 0.1003 & 0.4719 & 19.04 \\
Flux Kontext  \cite{labs2025flux1kontextflowmatching}   & 76.26 & 94.80 & 78.86 & 51.26 & 92.58 & 50.05 & 0.5719 & 0.1068 & 0.5402 & 19.21 \\
FreeFine     \cite{zhu2025training}    & 90.26 & 96.09 & 91.33 & 82.11 & 94.91 & 71.30 & 0.1611 & 0.0564 & 0.2421 & 22.81 \\
Object Mover  \cite{yu2025objectmover}   & 85.32 & 96.61 & 88.61 & 80.08 & 96.21 & 77.46 & 0.1849 & {0.0449} & 0.1895 & {24.06} \\
\hline
Ours &  91.74 & \textbf{97.59} & \textbf{94.67} & \textbf{87.40} & \textbf{97.57} & \textbf{87.37}  & \textbf{0.1180} & \textbf{0.0292} & \textbf{0.1012}  & \textbf{24.97} \\
\bottomrule
\end{tabular}%
}
\end{table*}

\subsection{Comparisons with Competing Methods}

\begin{figure}
    \centering
    \includegraphics[width=1.0\linewidth]{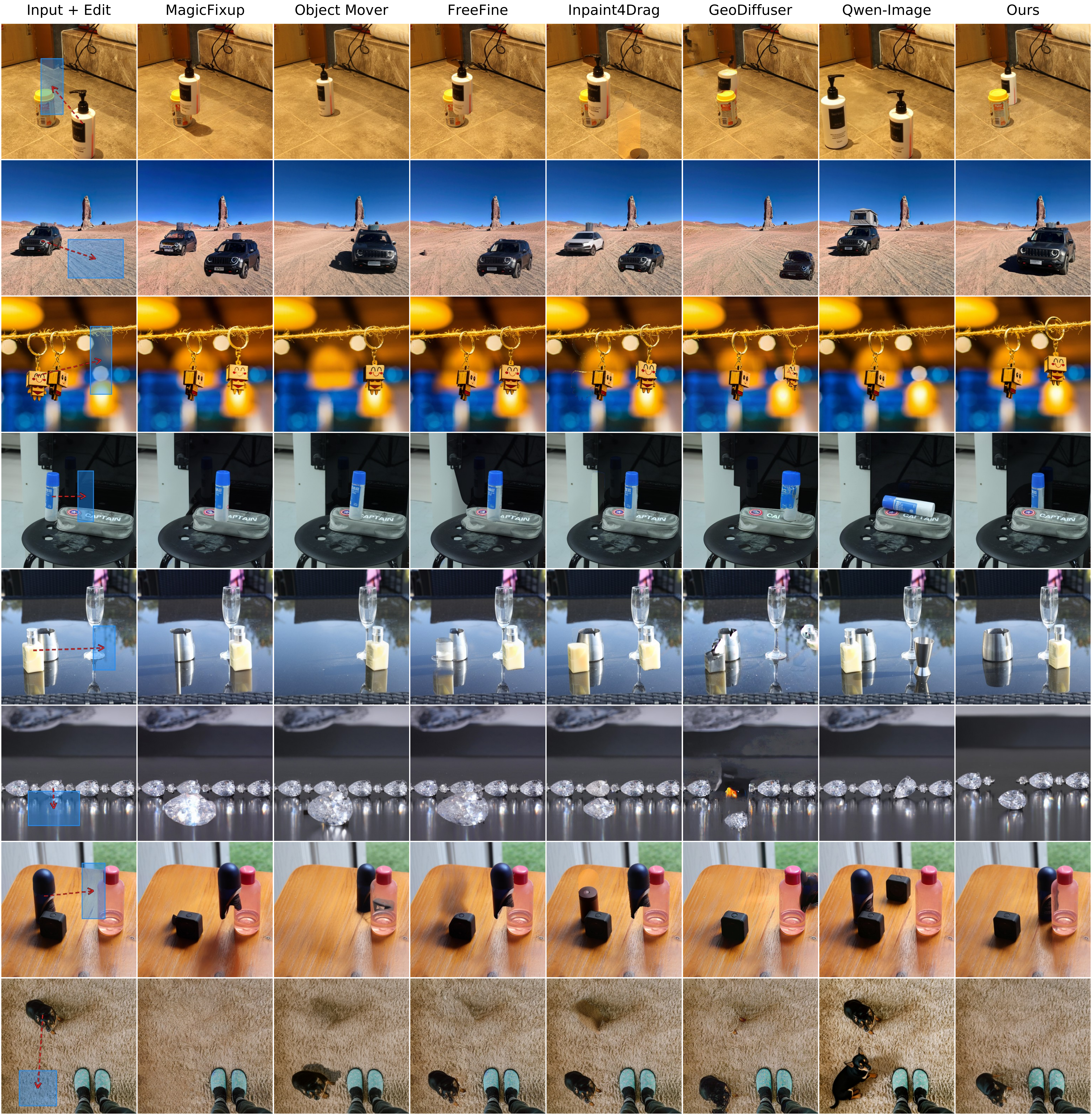}
    \caption{Qualitative comparisons with competing methods. }
    \label{fig:benchmarkA}
\end{figure}

We present quantitative and qualitative comparisons in Table~\ref{tab:comparison_test_split} and Fig.~\ref{fig:benchmarkA}, respectively. While we conduct an extensive evaluation against a broad set of baselines, we report only the strongest competing methods in the main paper and defer additional results to the supplementary material.

As shown in Fig.~\ref{fig:benchmarkA}, two-stage dragging and relocation methods such as MagicFixUp \cite{alzayer2025magic}, FreeFine \cite{zhu2025training}, and Inpaint4Drag \cite{lu2025inpaint4drag} struggle to model scene-dependent interactions, including shadows, reflections, and indirect illumination. Consequently, edited objects often appear physically inconsistent (e.g., floating or poorly grounded), which is also reflected in their quantitative performance. Although Inpaint4Drag achieves a relatively high CLIP-Score (target), its outputs largely resemble unrealistic copy-paste operations, as illustrated in the figure.

GeoDiffuser~\cite{sajnani2025geodiffuser}, a zero-shot optimization-based approach that incorporates geometric transformations within attention layers, requires inversion and exhibits poor object identity preservation. The base Qwen-Image model often fails to reliably relocate objects. The strongest competing method, ObjectMover~\cite{yu2025objectmover}, frequently removes the target object instead of relocating it, or produces duplicates (e.g., in the diamond example).

In contrast, our method consistently handles challenging scenarios involving occlusion and complex scene interactions. For example, it correctly relocates objects behind transparent surfaces (e.g., placing the deodorant behind the bottle while preserving transparency effects), accurately moves small and structured objects such as the diamond. Quantitatively, our approach achieves significant improvements across all evaluation metrics. Our method also enables object addition by realistically blending copy-pasted objects into the input scene and generalizes to other DiT-based editing models such as FLUX.1 Kontext (See Fig. \ref{fig:image_enhancement}, \ref{fig:flux} in supplementary material).

\subsection{Ablation Study}

\begin{figure}
    \centering
    \includegraphics[width=1.0\linewidth]{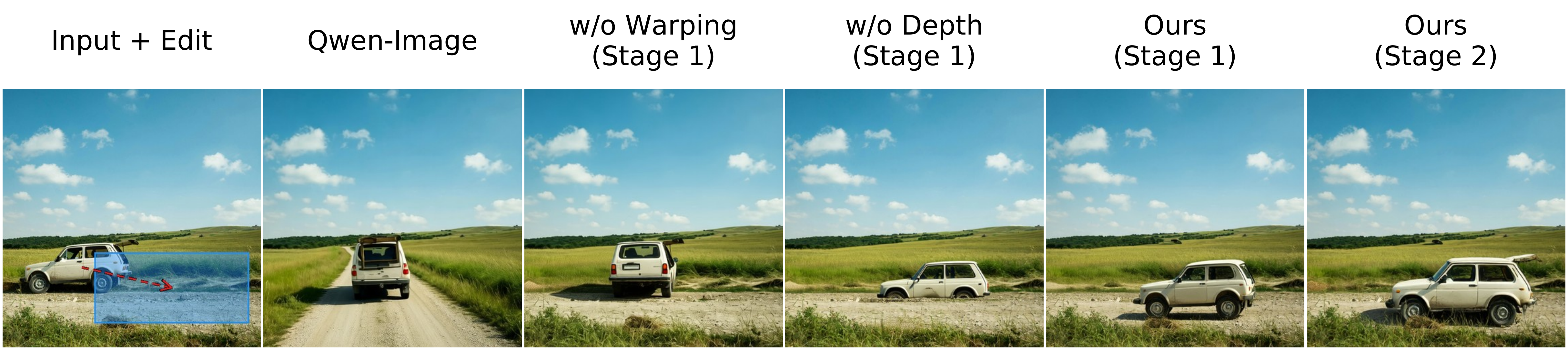}
    \caption{Ablation on training setup}
    \label{fig:training_ablation_main}
\end{figure}

Fig.~\ref{fig:training_ablation_main} presents an ablation study of the architectural modifications introduced in our method. 
We begin with a pretrained Qwen-Image editing model, which is capable of performing a variety of edits; however, it lacks precision for object relocation, and text prompts alone do not allow for accurate spatial control.
Next, we fine-tune the model using LoRA on our datasets without any additional modifications which we denote as \emph{w/o Warping (Stage 1)}. 
While this improves edit consistency, precise control over object motion remains limited.

Introducing RoPE-based spatial warping (\emph{w/o Depth (Stage 1)}) leads to a significant improvement, enabling accurate control over the horizontal and vertical $(x, y)$ position of the object. 
However, without incorporating the depth information, the resulting object placement can still deviate from the intended spatial configuration, particularly along the depth axis.
By augmenting the RoPE-based warping with depth information, we achieve more accurate object placement, resolving ambiguities along the depth axis and improving overall spatial consistency.
We further compare models trained only on the synthetic CLEVR dataset (\emph{Stage 1}) with the final model obtained after additional fine-tuning on real data (\emph{Stage 2}). 
While the Stage 1 model successfully learns to relocate objects, it often struggles to preserve fine-grained appearance details. 
In contrast, the second stage substantially improves identity preservation and visual fidelity, demonstrating the benefit of even a small amount of real-image supervision. Additional ablation studies are provided in Section \ref{subsec:supp_ablation}.

\begin{figure}
    \centering
    \includegraphics[width=1.0\linewidth]{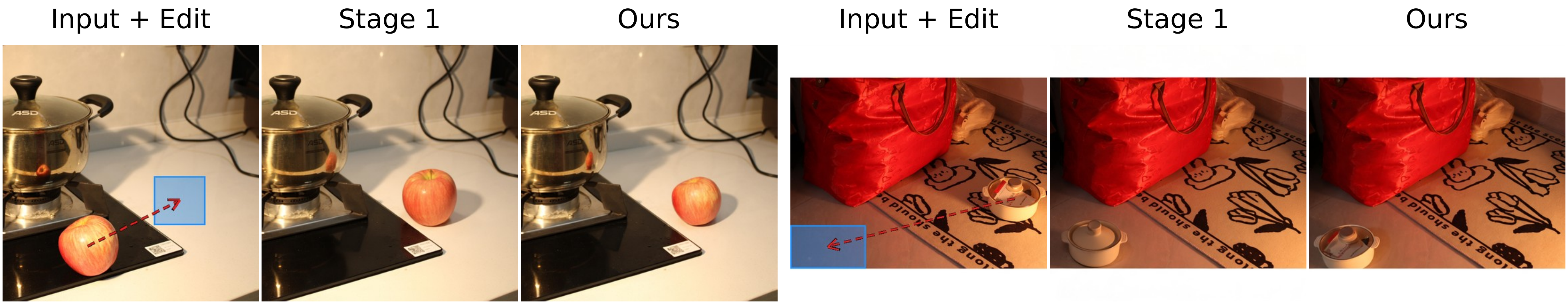}
    \caption{Identity improvement with second stage training on real captured dataset.}
    \label{fig:2stage_main}
\end{figure}

Fig.~\ref{fig:2stage_main} provides additional qualitative comparisons between the model trained only on synthetic data (Stage 1) and the final model after real-data fine-tuning (Stage 2). 
Consistent with the previous observations, the Stage 1 model can reliably relocate objects but often fails to preserve fine-grained appearance details. 
The second stage substantially improves identity preservation and overall visual fidelity.
Notably, the model also captures secondary effects such as reflections: in the first example, the reflection of the apple on the pot is coherently relocated along with the object, indicating that the learned transformation extends beyond the object itself to its visual interactions with the scene.
\section{Conclusion}\label{sec:conclusion}

We presented a geometry-aware object relocation method that operates directly on the internal representations of diffusion transformers. By interpreting rotary positional embeddings (RoPE) as a manipulable spatial field and extending them with depth-aware structure, our approach enables controlled object relocation with consistent handling of occlusions and scene-dependent effects.  Extensive experiments demonstrate strong improvements in geometric consistency, visual fidelity, and localization accuracy over prior methods. 

\textbf{Limitations.} Our method inherits the computational cost of multi-step diffusion sampling, resulting in non-trivial inference time, with negligible overhead from depth-preprocessing. 
In addition, since our approach builds on pre-trained generative models, it may inherit their biases and failure modes.

\textbf{Broader Impact.}
Our work enables more controllable and geometry-aware image editing, which can benefit applications in content creation, visual effects, and interactive design by reducing manual effort and improving realism. 
At the same time, such technology can be misused to create realistic but manipulated images, potentially contributing to misinformation or deceptive media. While our method focuses on geometric consistency rather than semantic manipulation, it still lowers the barrier to producing visually convincing edits. We therefore encourage the use of appropriate safeguards, such as watermarking, provenance tracking, and responsible deployment practices.

\bibliographystyle{plainnat}
\bibliography{references}

\clearpage
\appendix

\section{Technical Appendices and Supplementary Material}

This supplementary material provides additional details and results that complement the main paper. Section~\ref{subsec:supp_datasets} describes our synthetic and real training datasets in detail. Section~\ref{subsec:supp_training} reports training hyperparameters and setup. Section~\ref{subsec:supp_ablation} presents extended ablation studies on training data scale, architectural components, and the inference-time warping schedule. Section~\ref{subsec:supp_qualitative} provides additional qualitative comparisons against competing methods. Section~\ref{subsec:supp_generalization} demonstrates the generalization of our method to other DiT-based editing models and to an object addition application. Finally, Section~\ref{subsec:user_study} reports the results of our user study.

\subsection{Training Datasets}\label{subsec:supp_datasets}
\paragraph{Synthetic Data Generation with Clevr.}
We rely on synthetic data because it provides controllable ground-truth drag offsets and depth maps together with diverse object shapes, sizes, materials, and colors. Capturing such controlled variation is difficult to obtain at scale from natural footage. We extend the CLEVR \cite{johnson2016clevrdiagnosticdatasetcompositional} scene generator into a paired-rendering pipeline: each scene of two to six primitives is rendered, one object is then translated by a randomly sampled 3D vector subject to ground, spacing, frustum, and non-intersection constraints, and the scene is re-rendered with identical camera, lighting, and materials. This yields a pixel-accurate before/after pair where the only change is the moved object and the disocclusion it reveals. Each pair is stored together with the visible and fully-unoccluded masks of the moved object, per-frame depth maps, and the 3D move information used to derive the natural-language drag prompt. The drag inputs $(\Delta x, \Delta y)$ are taken directly from the projected pixel displacement of the moved object recorded in the renderer.

%, while the depth offset $\Delta z$ is computed as the difference between the mean MoGe-2~\cite{wang2025moge2accuratemonoculargeometry} depths under the source and target object masks \ceylan{do we need to specify the depth offset for training since we don't use it?}.

The pipeline is implemented in Blender~3.6 \cite{blender3.6} with the Cycles renderer. All quantities below are expressed in Blender world units; the orthographic camera (\texttt{ortho\_scale}=8.0) views a square ground region inside which objects are placed at $(x, y) \in [-3, 3]^2$, with object radii of $0.35$ and $0.70$ for small and large primitives, respectively. Each scene contains $N \in \{2, \ldots, 6\}$ primitives from CLEVR's catalogue of shapes, sizes, materials, and colors. Candidate positions are accepted only if the gap to every existing object satisfies $\mathrm{dist} - r_i - r_j \geq 0.25$ and the per-axis margin against each cardinal direction exceeds $0.4$; the scene is re-rolled if any object's visible footprint (verified via a shadeless flat-color render) falls below $200$ pixels. The source frame is rendered at $720\times480$ with $512$ Cycles samples and $8$ light bounces under three-point lighting jittered by $1.0$ unit per lamp, and the camera is jittered by $0.5$ units. %\ceylan{is the scene normalized, if so mention it so the distance values make sense}

For simulating object movements, in each scene, one object is selected uniformly at random and translated by a vector sampled from a randomly chosen axis subset $\mathcal{A} \in \{x, y, z, xy, xz, yz, xyz\}$ and scaled to magnitude $d \sim \mathcal{U}[1, 12]$; per-axis components are drawn from $\mathcal{U}[-1, 1]$ except $\Delta z \sim \mathcal{U}[0, 1]$ to bias toward upward motion. A candidate is accepted only if it (i) stays above the ground, (ii) preserves inter-object spacing and cardinal margins, (iii) keeps $\geq 50\%$ of its projected bounding box inside the frustum, (iv) has no mesh intersection with any other object (\texttt{BVHTree.overlap} on the evaluated depsgraph, with static BVHs prebuilt once), and (v) re-passes the per-object visibility test. Up to $30$ candidates are tried before the sample is discarded. The scene is then re-rendered with identical camera, lighting, and material state.

Each pair is saved with (a) the \emph{visible} object mask; (b) the \emph{fully-unoccluded} mask, obtained by hiding all other objects, replacing the target's material with a white emission shader, and rendering at one Cycles sample on a black background; (c) per-image depth maps for source and target frames from MoGe-2~\cite{wang2025moge2accuratemonoculargeometry}; (d) a \texttt{move\_info} record with the object index, old and new 3D coordinates, signed per-axis deltas $(\Delta x, \Delta y, \Delta z)$, move distance, axis-subset label, and a textual direction descriptor (combinations of \emph{left/right}, \emph{forward/backward}, \emph{up/down}) derived by projecting $\boldsymbol{\Delta}$ onto the camera-relative basis. The descriptor is used to construct the natural-language drag prompt.

%per-frame depth maps are extracted from \cite{moge}
%per-frame Z-pass depth maps written as multilayer EXRs through a compositor node graph; and (d) 

\begin{figure}
    \centering
    \includegraphics[width=1.0\linewidth]{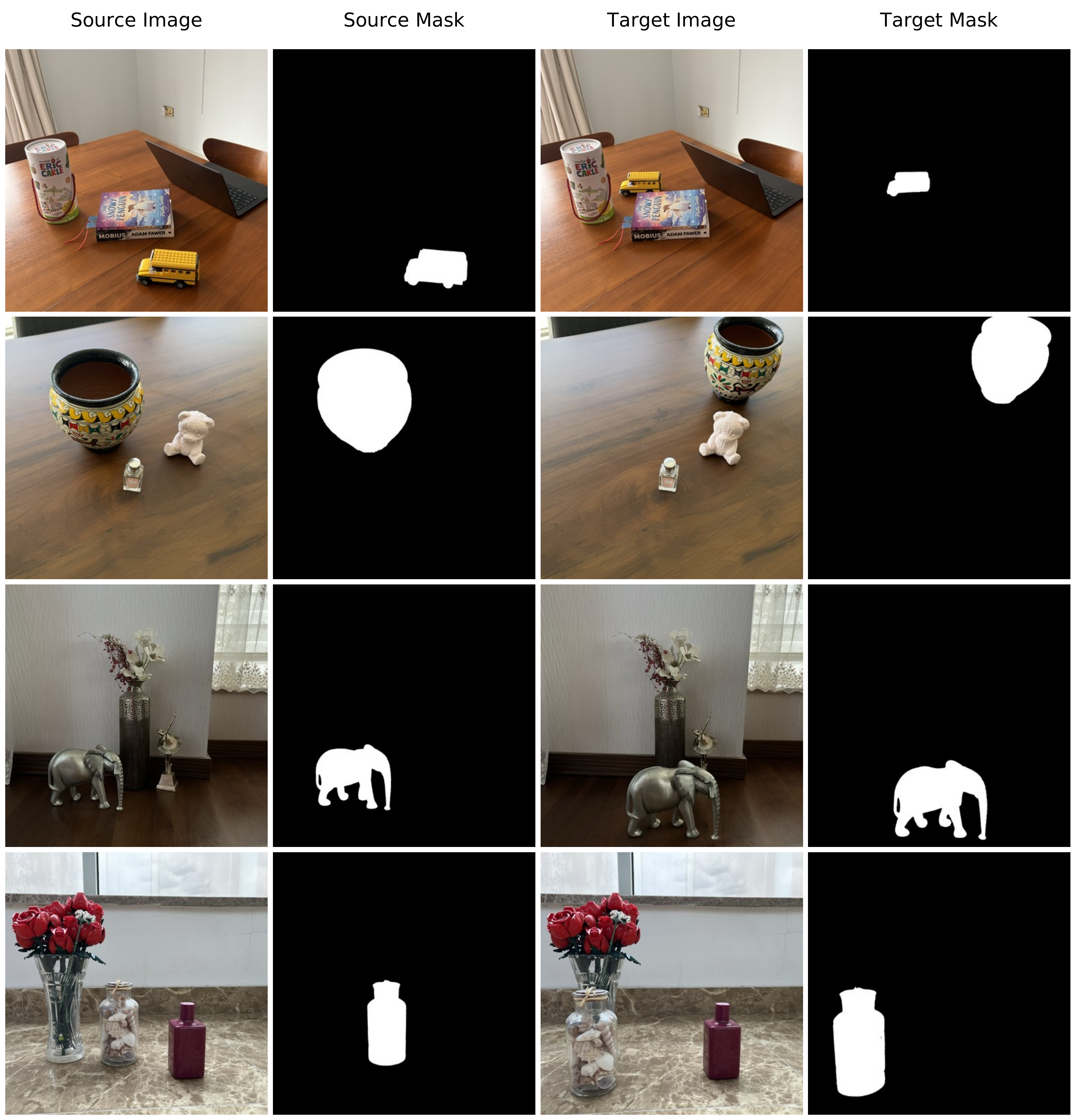}
    \caption{Examples from our captured real-photo dataset. Each row shows the source image, the source-object mask, the target image after the drag, and the target mask.}
    \label{fig:captured_example}
\end{figure}

\paragraph{Real Data Collection}
For the second-stage training, we fine-tune our model on a dataset of real photographs. In a controlled setup, we captured 166 paired examples: for each scene we record the source image with the object at one location and the target image with the same object moved to a second location, together with a clean background frame of the empty scene. Both frames are captured from the same viewpoint under matched lighting, so the only RGB change between source and target is the relocated object and the disocclusion it reveals. Object masks for both frames are produced with SAM~\cite{sam}, and the natural-language drag prompt is generated by Qwen3-VL~\cite{qwen3-vl} from the source/target image pair, since no structured move information is available for real captures. The pixel-space drag $(\Delta x, \Delta y)$ is recovered as the displacement between the centroids of the source and target object masks, the depth offset $\Delta z$ is obtained as the difference between the mean MoGe-2~\cite{wang2025moge2accuratemonoculargeometry} depths under the two masks, and an apparent-size scale factor is derived from the square root of the target/source mask-area ratio. Representative examples from this dataset are shown in Fig.~\ref{fig:captured_example}.

% LoRA fine-tuning on DiT (rank 16, target modules)
% ZeRO-2, 4 GPUs, dataset repeat, loss objective
% Training on CLEVR-2k + captured (real dataset)

\subsection{Training Details}\label{subsec:supp_training}
We fine-tune the Qwen-Image-Edit-2511 transformer with rank-16 LoRA adapters injected into the attention projection layers (\texttt{to\_q}, \texttt{to\_k}, \texttt{to\_v}, \texttt{add\_q\_proj}, \texttt{add\_k\_proj}, \texttt{add\_v\_proj}, \texttt{to\_out.0}, \texttt{to\_add\_out}), the MLP projections (\texttt{img\_mlp.net.2}, \texttt{txt\_mlp.net.2}), and the modulation layers (\texttt{img\_mod.1}, \texttt{txt\_mod.1}), while keeping the text encoder and VAE frozen. Training is performed on 4 NVIDIA A100 GPUs at $512\times512$ resolution in BF16 precision using DeepSpeed ZeRO-2, with gradient checkpointing and optimizer offload to fit the model in memory. We use the AdamW optimizer with a constant learning rate of $1\times10^{-4}$ and a per-GPU batch size of $1$ (effective batch size $4$).

We adopt a two-stage curriculum. In Stage~1, the model is trained on the synthetic CLEVR dataset for two epochs of $5{,}000$ optimizer steps each. In Stage~2, we initialize from the Stage~1 checkpoint and fine-tune on the captured real-photo dataset for two further epochs, with a dataset repeat factor of $30$ to compensate for its smaller size, bridging the synthetic-to-real gap. Stage~1 takes approximately $20$ hours per epoch, and Stage~2 takes approximately $8$ hours in total.

\subsection{Additional Ablation Studies}\label{subsec:supp_ablation}

\begin{figure}
    \centering
    \includegraphics[width=1.0\linewidth]{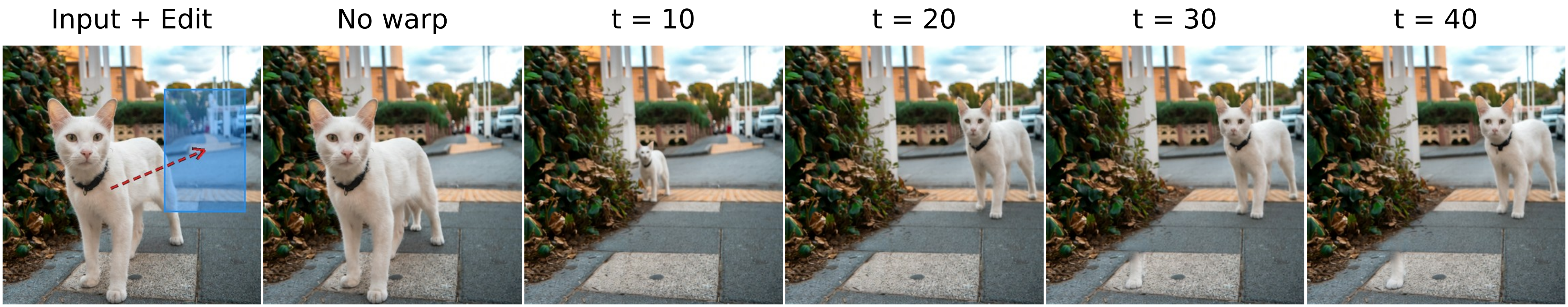}
    \caption{Effect of the warping schedule. Warping is applied for the first $t$ denoising steps. We use $t = 20$ in our experiments.}
    \label{fig:timestep}
\end{figure}

We complement the ablations in the main paper with four additional studies: the effect of synthetic dataset scale on Stage~1 (Table~\ref{tab:stage1_clevr_lora}), the effect of Stage~1 initialization on the Stage~2 model (Table~\ref{tab:stage2_captured_revised}), a quantitative breakdown of the contribution of each architectural component (Table~\ref{tab:ablation_warping}), and the effect of the inference-time warping schedule (Fig.~\ref{fig:timestep}).

\paragraph{Stage 1: synthetic dataset scale.}
We evaluate Stage 1 models trained only on the synthetic CLEVR dataset at three scales (2k, 5k, and 10k paired scenes) to assess the effect of training data quantity, and report our final Stage 2 model in the last row of Table~\ref{tab:stage1_clevr_lora} for reference. As shown in Table~\ref{tab:stage1_clevr_lora}, all metrics generally improve as the synthetic dataset grows from 2k to 10k paired scenes, indicating that Stage 1 benefits from larger-scale synthetic supervision. Nonetheless, a clear gap remains between the best Stage 1 variant and our final Stage 2 model, highlighting the complementary role of real-image fine-tuning in closing the synthetic-to-real gap.

\begin{table*}[t]
\centering
\caption{\textbf{Stage 1 ablation: training dataset size on synthetic data.} Evaluation is performed on ObjMove-A at $512{\times}512$. We report CLIP/DINO $\uparrow$, DreamSim $\downarrow$, PSNR $\uparrow$ on the target object (Tgt), source region (Src), and background (BG). \textbf{Bold} indicates the best score per column among Stage 1 variants.}
\label{tab:stage1_clevr_lora}
\resizebox{\textwidth}{!}{%
\begin{tabular}{l ccc ccc ccc c}
\toprule
 & \multicolumn{3}{c}{\textbf{CLIP Score} $\uparrow$} 
 & \multicolumn{3}{c}{\textbf{DINO Score} $\uparrow$}
 & \multicolumn{3}{c}{\textbf{DreamSim} $\downarrow$}
 & \textbf{PSNR} $\uparrow$ \\
\cmidrule(lr){2-4}\cmidrule(lr){5-7}\cmidrule(lr){8-10}
\textbf{Size} & Tgt & BG & Src & Tgt & BG & Src & Tgt & BG & Src & (dB) \\
\midrule
2k  & 88.38 & 96.45 & 93.30 & 79.52 & 95.62 & 81.93 & 0.21 & 0.05 & 0.15 & 22.94 \\
5k  & 88.23 & 96.49 & 93.62 & 80.49 & 95.81 & 83.49 & 0.20 & 0.05 & 0.13 & 22.45 \\
10k & \textbf{89.87} & \textbf{96.92} & \textbf{93.65} & \textbf{82.57} & \textbf{96.38} & \textbf{84.03} & \textbf{0.17} & \textbf{0.04} & \textbf{0.13} & \textbf{23.66} \\
\midrule
\textit{Ours (Stage 2)} & \textbf{91.74} & \textbf{97.59} & \textbf{94.67} & \textbf{87.40} & \textbf{97.57} & \textbf{87.37} & \textbf{0.12} & \textbf{0.03} & \textbf{0.10} & \textbf{24.97} \\
\bottomrule
\end{tabular}%
}
\end{table*}

\paragraph{Stage 2: effect of Stage 1 initialization.}
We further fine-tune each Stage 1 checkpoint from Table~\ref{tab:stage1_clevr_lora} (2k, 5k, and 10k synthetic scales) on our captured real-image dataset to assess how the Stage 1 initialization affects the final Stage 2 model. As shown in Table~\ref{tab:stage2_captured_revised}, real-data fine-tuning yields consistent improvements over Stage 1 across all initializations, with the 5k-initialized variant offering the best overall trade-off across metrics, which we adopt as our final model. This also indicates that 5k synthetic scenes are sufficient to support effective Stage 2 fine-tuning, motivating our choice of this scale for the final model.

\begin{table*}[t]
\centering
\caption{\textbf{Stage 2 ablation: real-data fine-tuning across Stage 1 initializations.} Evaluation is performed on ObjMove-A at $512{\times}512$. We report CLIP/DINO $\uparrow$, DreamSim $\downarrow$, PSNR $\uparrow$ on the target object (Tgt), source region (Src), and background (BG). \textbf{Bold} indicates the best score per column.}
\label{tab:stage2_captured_revised}
\resizebox{\textwidth}{!}{%
\begin{tabular}{l ccc ccc ccc c}
\toprule
 & \multicolumn{3}{c}{\textbf{CLIP Score} $\uparrow$} 
 & \multicolumn{3}{c}{\textbf{DINO Score} $\uparrow$}
 & \multicolumn{3}{c}{\textbf{DreamSim} $\downarrow$}
 & \textbf{PSNR} $\uparrow$ \\
\cmidrule(lr){2-4}\cmidrule(lr){5-7}\cmidrule(lr){8-10}
\textbf{Stage 1 Size} & Tgt & BG & Src & Tgt & BG & Src & Tgt & BG & Src & (dB) \\
\midrule
2k  & 91.71 & 97.50 & 94.61 & \textbf{87.59} & 97.58 & 87.36 & \textbf{0.12} & \textbf{0.03} & \textbf{0.10} & 24.99 \\
\textit{5k (Ours)}  & \textbf{91.74} & \textbf{97.59} & \textbf{94.67} & 87.40 & 97.57 & \textbf{87.37} & \textbf{0.12} & \textbf{0.03} & \textbf{0.10} & 24.97 \\
10k & 91.26 & 97.56 & 94.46 & 87.06 & \textbf{97.60} & 86.91 & \textbf{0.12} & \textbf{0.03} & 0.11 & \textbf{25.20} \\
\bottomrule
\end{tabular}%
}
\end{table*}

\paragraph{Architectural components.}
As shown in Table~\ref{tab:ablation_warping}, we report three Stage 1 variants of our method: \emph{w/o Warping (Stage 1)} denotes LoRA fine-tuning without RoPE-based spatial warping, \emph{w/o Depth (Stage 1)} adds warping without depth conditioning, and \emph{Ours (Stage 1)} is the full architecture without Stage 2 fine-tuning. Each variant is trained across multiple synthetic CLEVR scales (2k, 5k, 10k) as well as on our captured real dataset alone. Removing warping consistently degrades performance across all training scales, indicating that LoRA fine-tuning alone is insufficient for precise object relocation. Adding warping (\emph{w/o Depth (Stage 1)}) yields a marked improvement, and incorporating depth conditioning (\emph{Ours (Stage 1)}) further improves spatial accuracy. Finally, our full model with Stage 2 real-data fine-tuning, \emph{Ours (Stage 2)}, achieves the best results across all metrics. Qualitative results supporting these observations are provided in Fig.~\ref{fig:training_ablation_supp}.

\begin{table*}[t]
\centering
\caption{\textbf{Ablation on architectural modifications.} Evaluation is performed on ObjMove-A at $512{\times}512$. We report CLIP/DINO $\uparrow$, DreamSim $\downarrow$, PSNR $\uparrow$ on the target object (Tgt), source region (Src), and background (BG).}
\label{tab:ablation_warping}
\resizebox{\textwidth}{!}{%
\begin{tabular}{ll ccc ccc ccc c}
\toprule
 & & \multicolumn{3}{c}{\textbf{CLIP Score} $\uparrow$} 
 & \multicolumn{3}{c}{\textbf{DINO Score} $\uparrow$}
 & \multicolumn{3}{c}{\textbf{DreamSim} $\downarrow$}
 & \textbf{PSNR} $\uparrow$ \\
\cmidrule(lr){3-5}\cmidrule(lr){6-8}\cmidrule(lr){9-11}
\textbf{Method} & \textbf{Size} & Tgt & BG & Src & Tgt & BG & Src & Tgt & BG & Src & (dB) \\
\midrule
\multirow{4}{*}{\shortstack[l]{w/o Warping\\(Stage 1)}}
    & 2k       & 83.51 & 94.63 & 90.71 & 72.33 & 93.95 & 73.99 & 0.32 & 0.08 & 0.22 & 19.96 \\
    & 5k       & 83.35 & 94.92 & 89.03 & 71.85 & 94.54 & 71.14 & 0.32 & 0.07 & 0.26 & 20.26 \\
    & 10k      & 83.69 & 96.56 & 90.97 & 73.75 & 95.87 & 77.06 & 0.31 & 0.05 & 0.21 & 21.40 \\
    & Captured & 84.18 & 95.26 & 90.29 & 74.41 & 95.07 & 75.13 & 0.30 & 0.06 & 0.22 & 20.85 \\
\midrule
\multirow{4}{*}{\shortstack[l]{w/o Depth\\(Stage 1)}}
    & 2k       & 87.51 & 96.47 & 93.48 & 77.29 & 95.50 & 83.79 & 0.24 & 0.06 & 0.14 & 23.38 \\
    & 5k       & 88.84 & 96.13 & 92.17 & 79.66 & 96.37 & 80.55 & 0.18 & 0.04 & 0.18 & 24.54 \\
    & 10k      & 87.99 & 96.92 & 93.48 & 78.21 & 96.32 & 84.33 & 0.27 & 0.05 & 0.22 & 23.56 \\
    & Captured & 87.08 & 96.20 & 93.38 & 75.89 & 95.37 & 83.70 & 0.25 & 0.06 & 0.14 & 22.67 \\
\midrule
\multirow{4}{*}{\shortstack[l]{Ours\\(Stage 1)}}
    & 2k       & 88.38 & 96.45 & 93.30 & 79.52 & 95.62 & 81.93 & 0.21 & 0.05 & 0.15 & 22.94 \\
    & 5k       & 88.23 & 96.49 & 93.62 & 80.49 & 95.81 & 83.49 & 0.20 & 0.05 & 0.13 & 22.45 \\
    & 10k      & 89.87 & 96.92 & 93.65 & 82.57 & 96.38 & 84.03 & 0.17 & 0.04 & 0.13 & 23.66 \\
    & Captured & 89.45 & 96.59 & 93.66 & 82.62 & 96.40 & 84.17 & 0.17 & 0.05 & 0.14 & 22.77 \\
\midrule
\textit{Ours (Stage 2)} & --- & \textbf{91.74} & \textbf{97.59} & \textbf{94.67} & \textbf{87.40} & \textbf{97.57} & \textbf{87.37} & \textbf{0.12} & \textbf{0.03} & \textbf{0.10} & \textbf{24.97} \\
\bottomrule
\end{tabular}%
}
\end{table*}

\paragraph{Warping schedule.}
At inference, RoPE-based spatial warping is applied for the first $t$ denoising steps and disabled thereafter, allowing the remaining steps to refine appearance details. As shown in Fig.~\ref{fig:timestep}, applying warping for too few steps leaves the object insufficiently relocated, while applying it for too many steps degrades visual quality. We use $t=20$ in all of our experiments.

\subsection{Additional Qualitative Results}\label{subsec:supp_qualitative}
We provide extensive qualitative comparisons against the baselines reported in the main paper, as well as additional methods omitted from the main paper for space.

\begin{figure}
    \centering
    \includegraphics[width=1.0\linewidth]{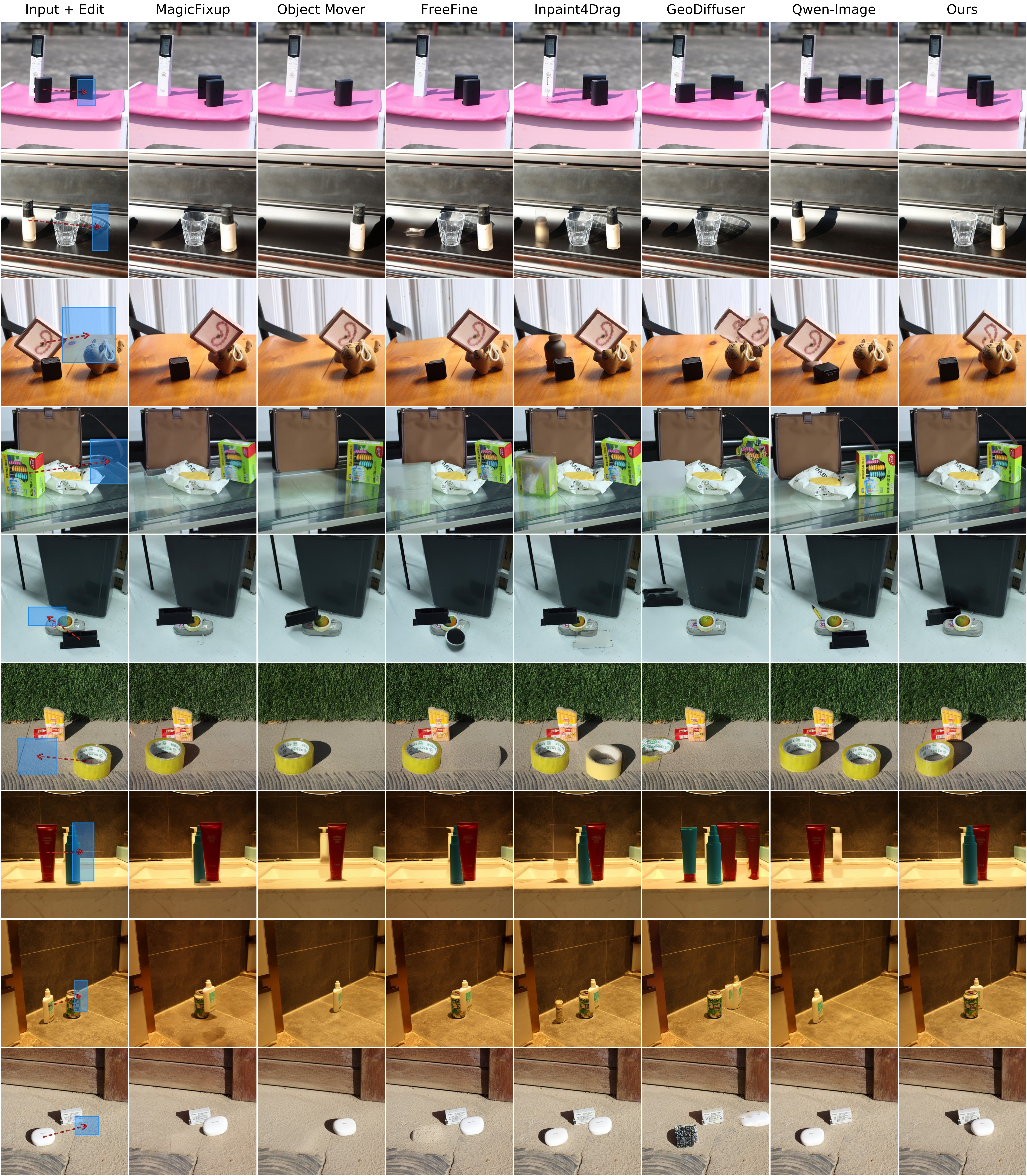}
    \caption{Additional qualitative comparisons on Object Mover Benchmark A.}
    \label{fig:supp_main_A}
\end{figure}

\begin{figure}
    \centering
    \includegraphics[width=0.95\linewidth]{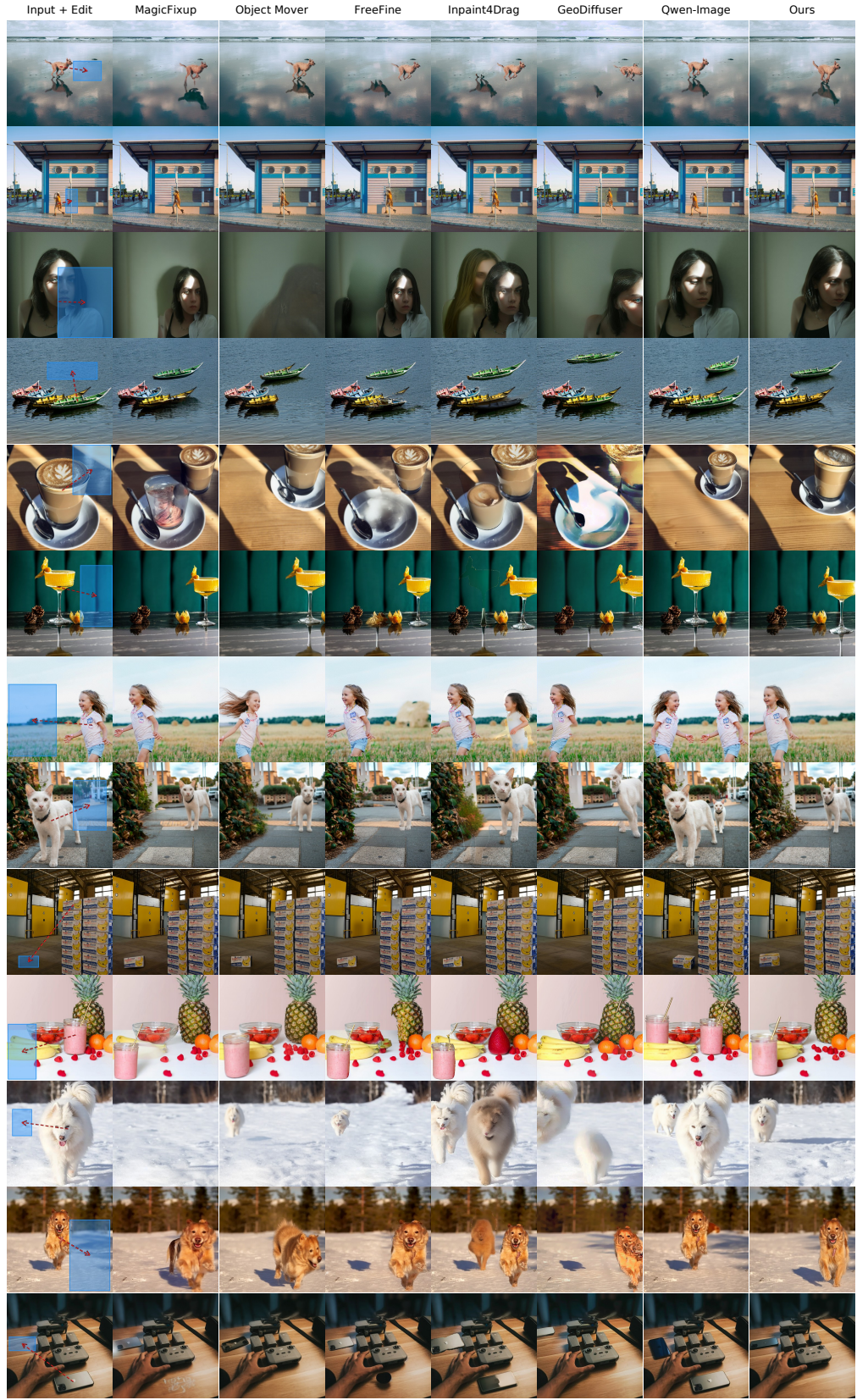}
    \caption{Additional qualitative comparisons on Object Mover Benchmark B.}
    \label{fig:supp_main_B}
\end{figure}

\begin{figure}
    \centering
    \includegraphics[width=0.9\linewidth]{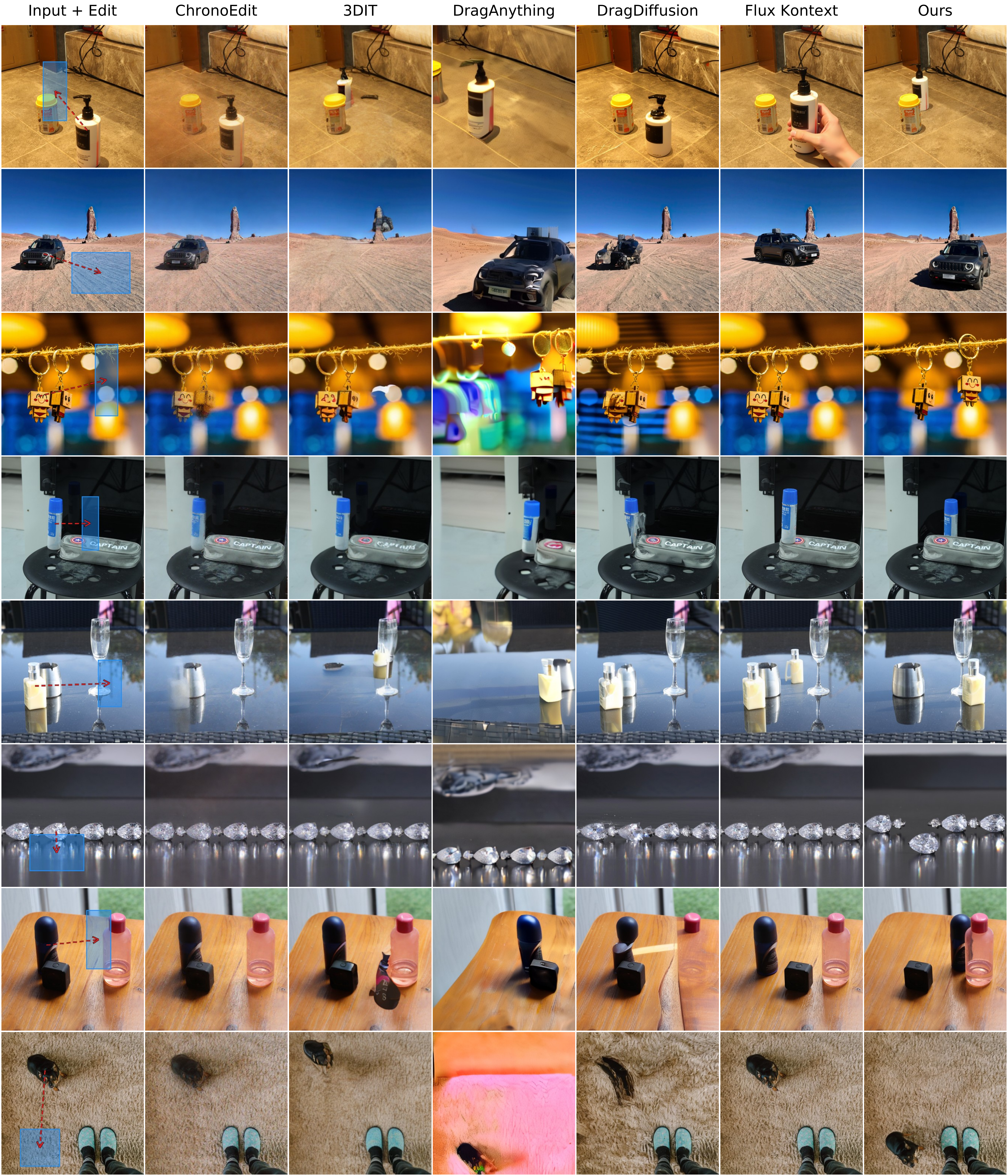}
    \caption{Qualitative comparisons against an extended set of methods on Object Mover Benchmarks A and B.}
    \label{fig:supp_mixed}
\end{figure}

\begin{figure}
    \centering
    \includegraphics[width=0.9\linewidth]{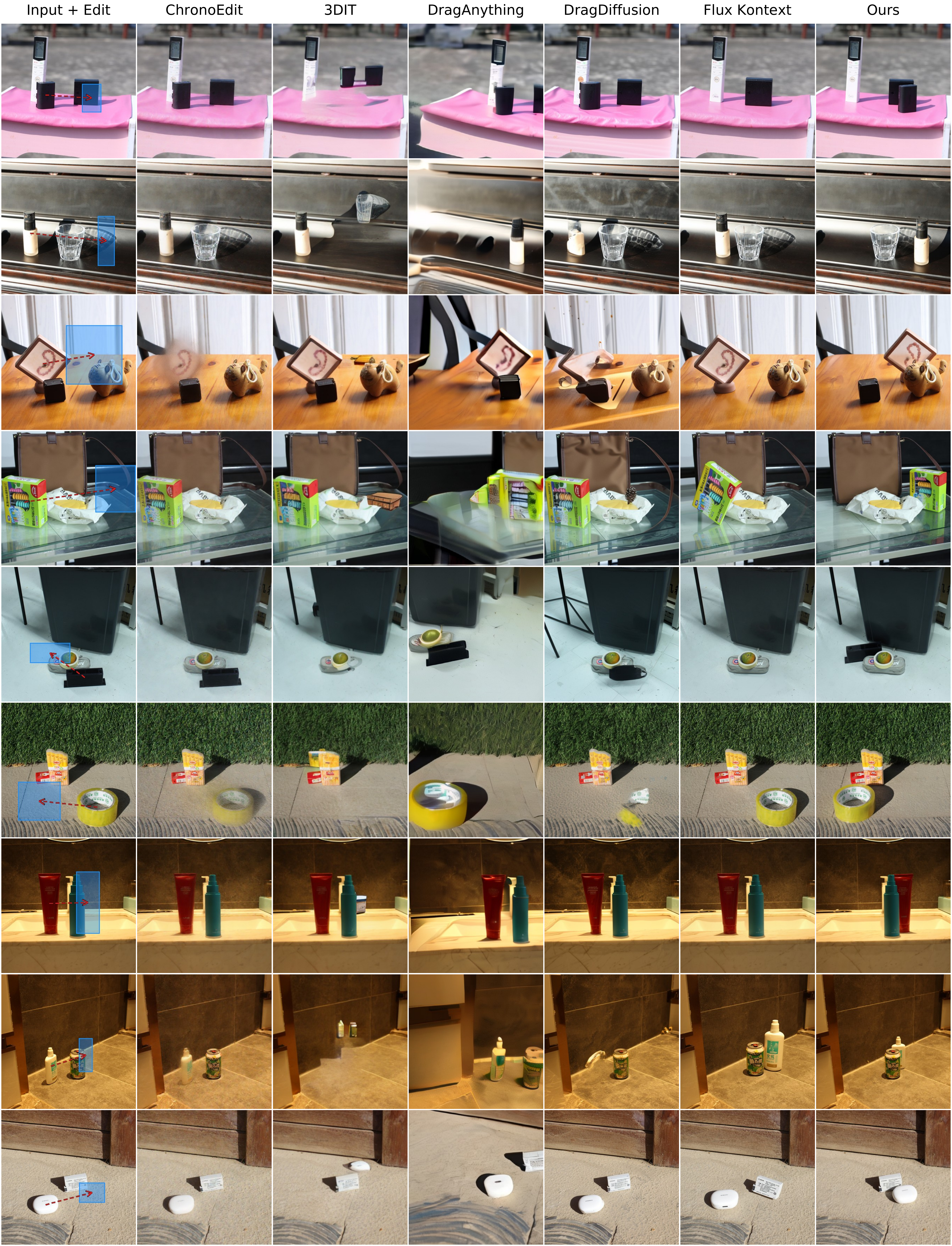}
    \caption{Qualitative comparisons against an extended set of methods on Object Mover Benchmarks A.}
    \label{fig:supp_extended_A}
\end{figure}

\begin{figure}
    \centering
    \includegraphics[width=0.82\linewidth]{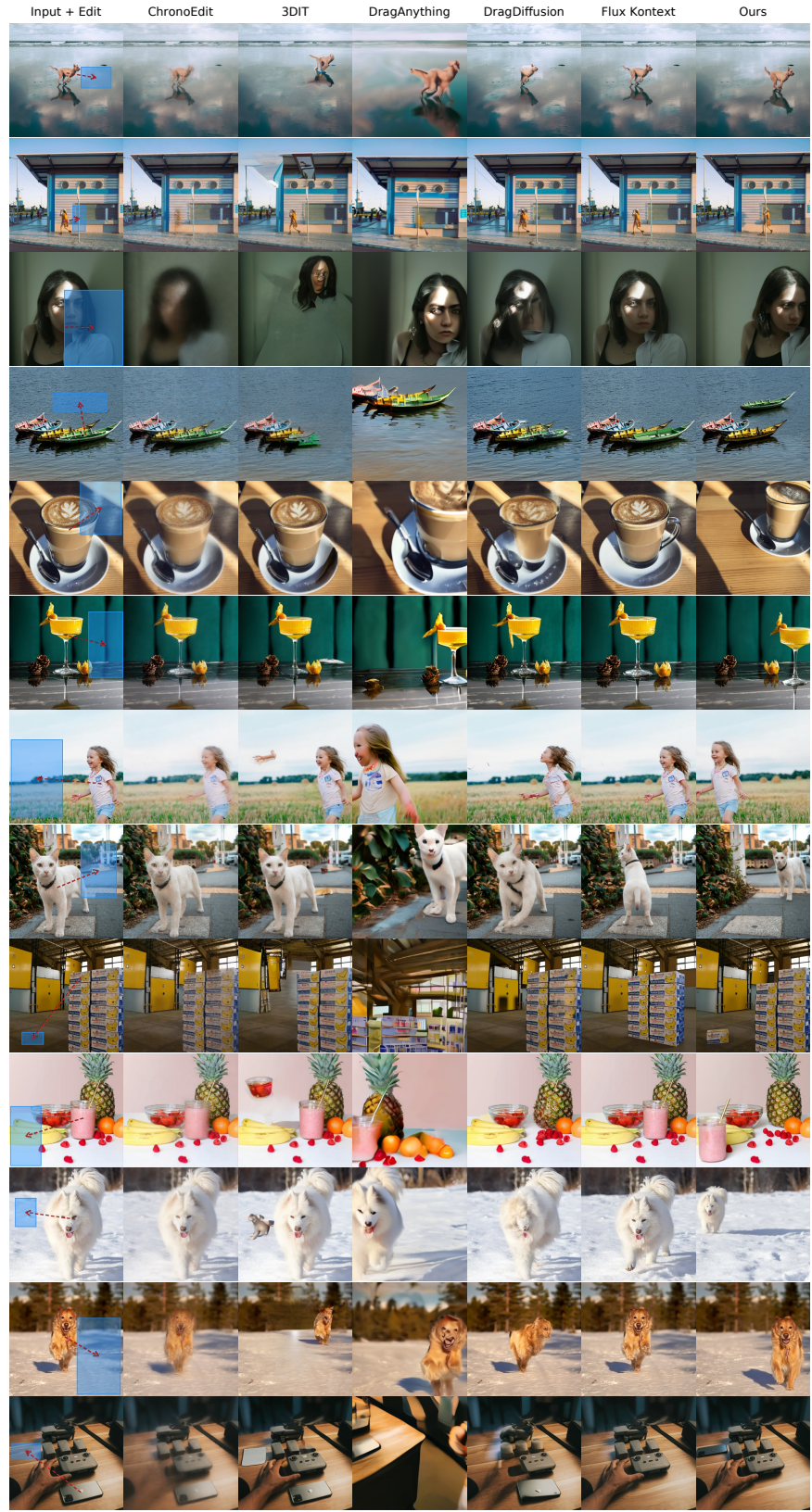}
    \caption{Qualitative comparisons against an extended set of methods on Object Mover Benchmarks B.}
    \label{fig:supp_extended_B}
\end{figure}

\begin{figure}
    \centering
    \includegraphics[width=1.0\linewidth]{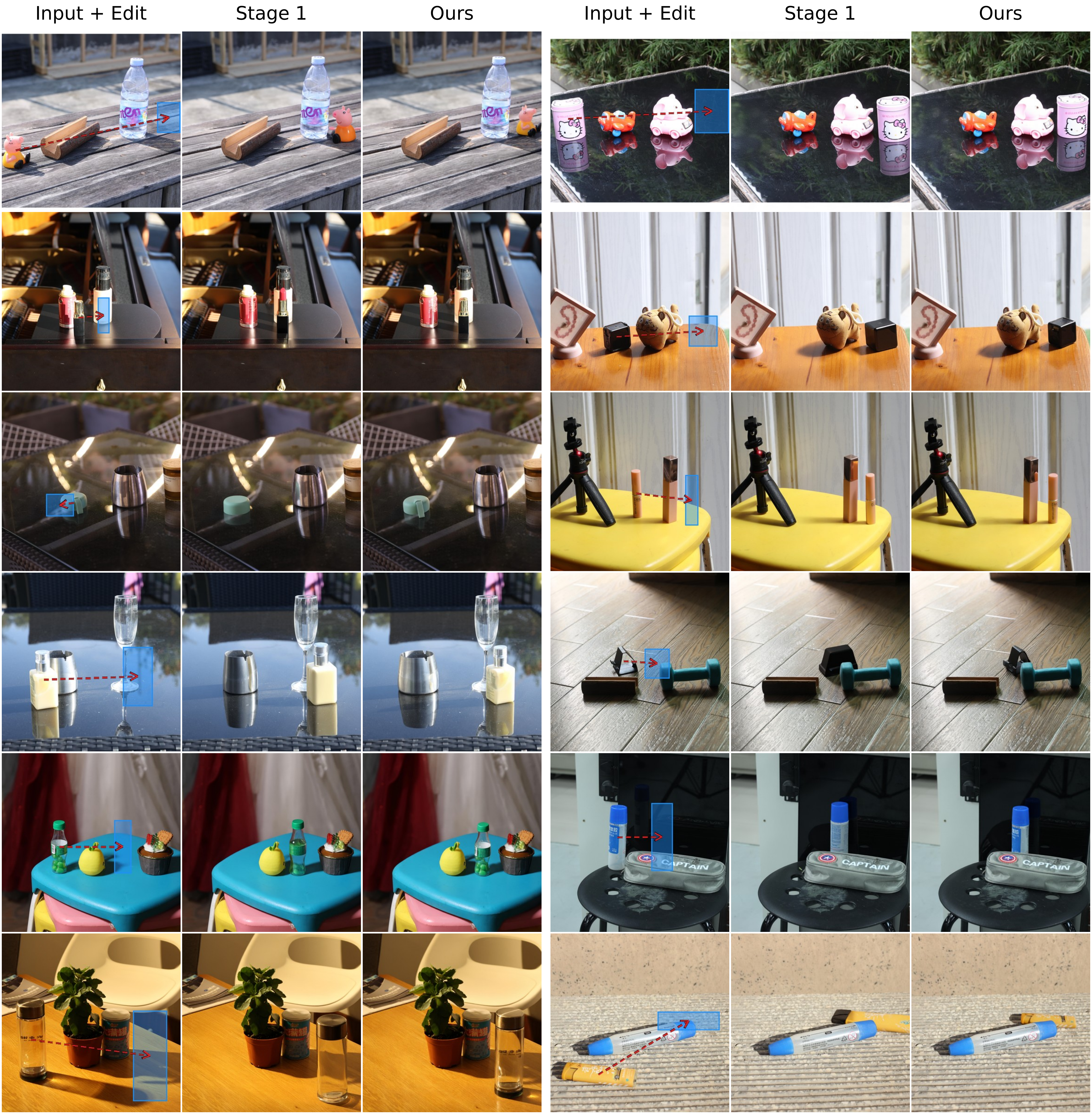}
    \caption{Additional results on two-stage training on Object Mover Benchmark A.}
    \label{fig:2stage_supp_A}
\end{figure}

\begin{figure}
    \centering
    \includegraphics[width=1.0\linewidth]{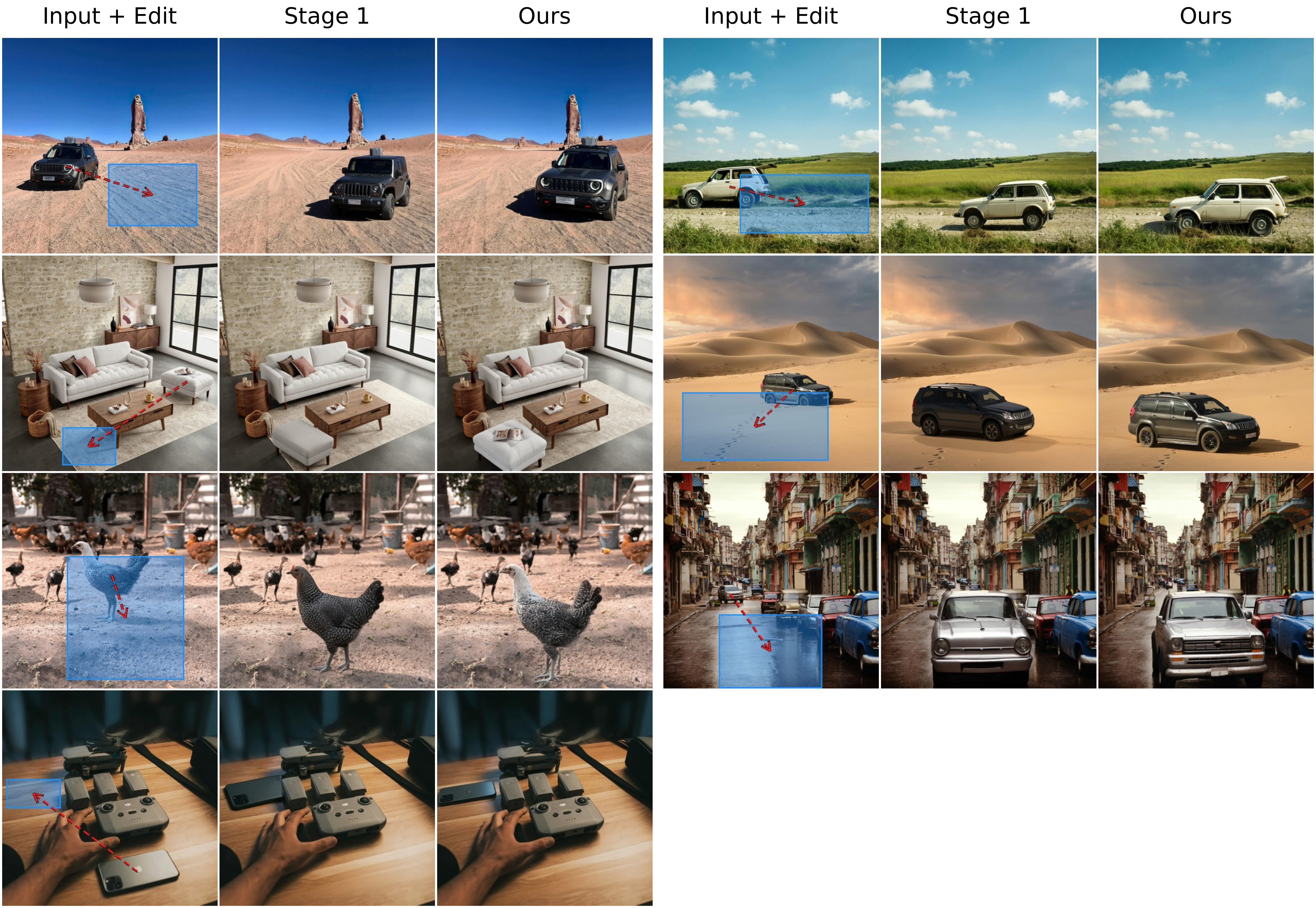}
    \caption{Additional results on two-stage training on Object Mover Benchmark B.}
    \label{fig:2stage_supp_B}
\end{figure}

\paragraph{Effect of two-stage training.}
Figs.~\ref{fig:2stage_supp_A} and~\ref{fig:2stage_supp_B} provide additional qualitative comparisons between the Stage 1 model (trained only on synthetic CLEVR data) and our final Stage 2 model (further fine-tuned on the captured real-photo dataset). Across both benchmarks, the Stage 2 model substantially improves identity preservation and visual fidelity, consistent with the trends reported in the main paper.

\begin{figure}
    \centering
    \includegraphics[width=1.0\linewidth]{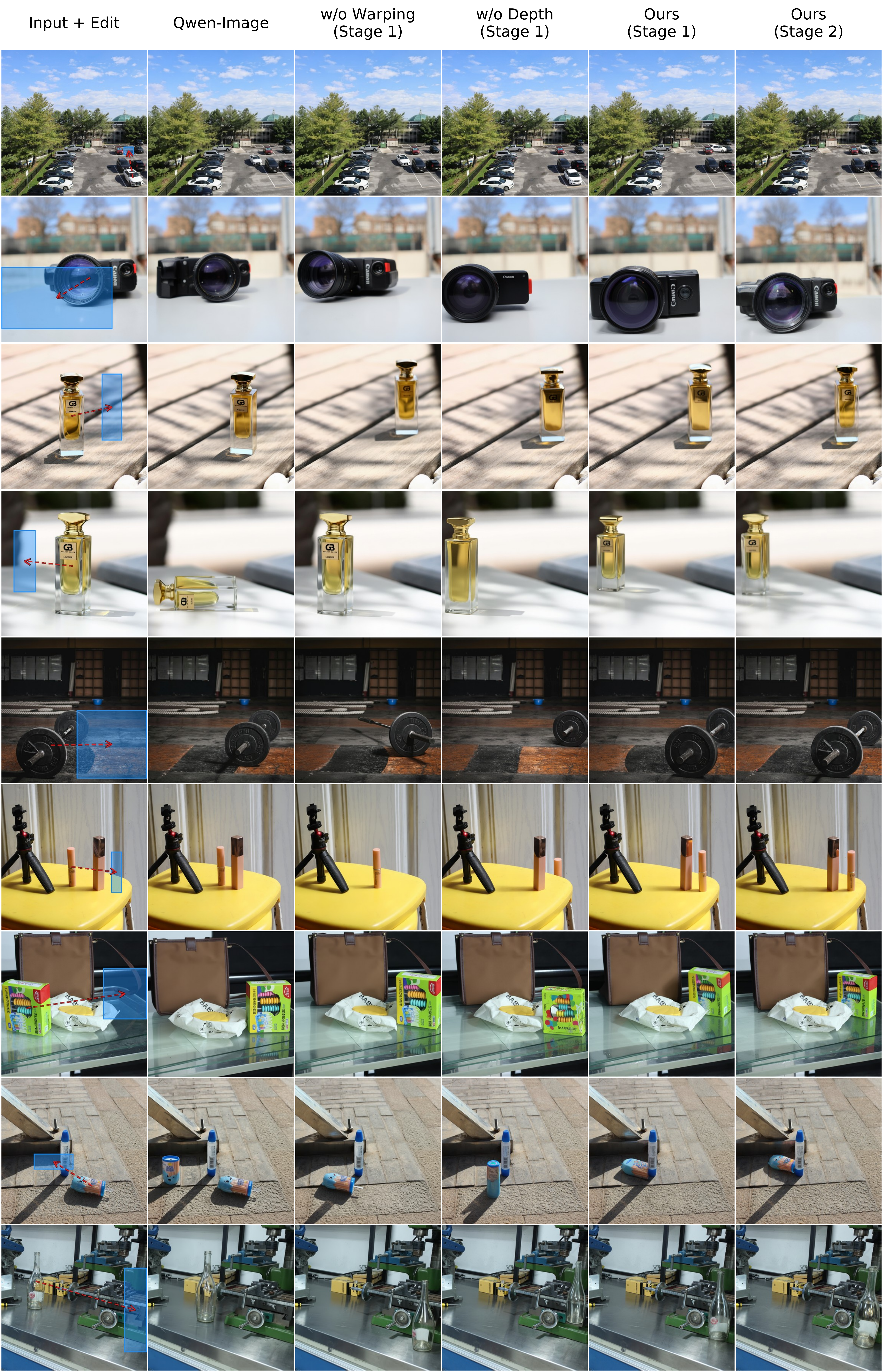}
    \caption{Component ablation across ObjMove-A and ObjMove-B. Removing depth or the second stage degrades quality; the full model is consistently best.}
    \label{fig:training_ablation_supp}
\end{figure}

\paragraph{Architectural component ablation.}
Fig.~\ref{fig:training_ablation_supp} provides qualitative results corresponding to the architectural ablation in Table~\ref{tab:ablation_warping}. Removing depth conditioning or the second-stage real-data fine-tuning visibly degrades quality, while the full model is consistently best across both benchmarks.

\begin{figure}
    \centering
    \includegraphics[width=1.0\linewidth]{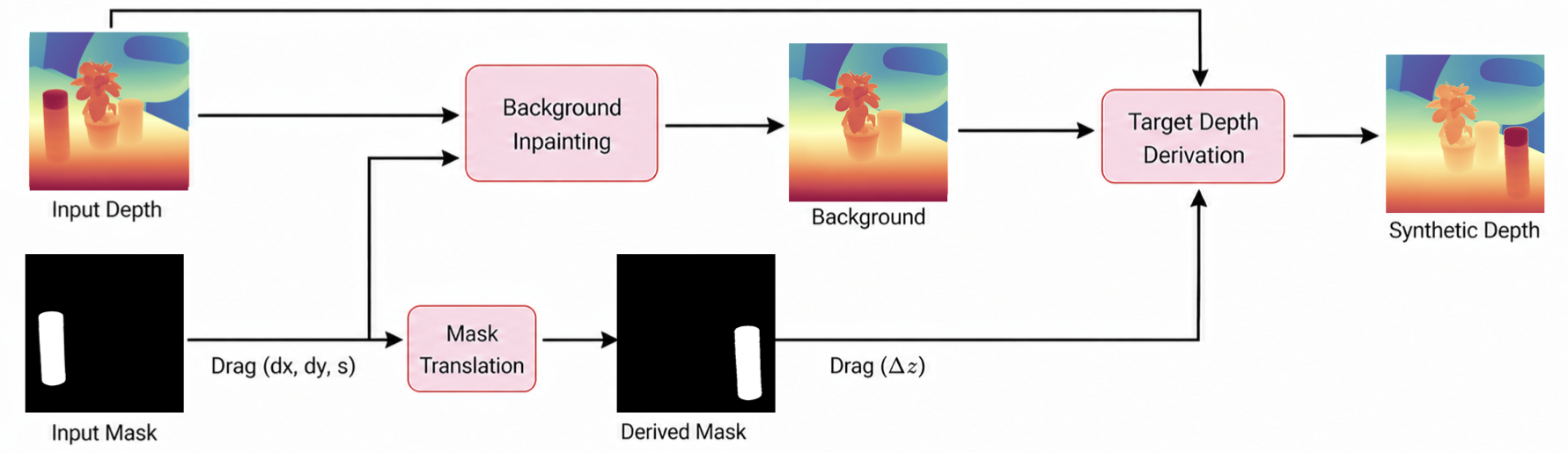}
    \caption{Target depth synthesis pipeline.}
    \label{fig:depth_synthesis}
\end{figure}

\paragraph{Target depth synthesis.}
Fig.~\ref{fig:depth_synthesis} illustrates our target depth synthesis pipeline, which combines Laplacian background fill, reverse warping, and Z-buffer compositing to produce a target depth map at inference time.

\subsection{Generalization and Applications}\label{subsec:supp_generalization}

Beyond object relocation on the ObjMove benchmarks, our method generalizes to other DiT-based editing backbones and supports related editing applications.

\begin{figure}
    \centering
    \includegraphics[width=1.0\linewidth]{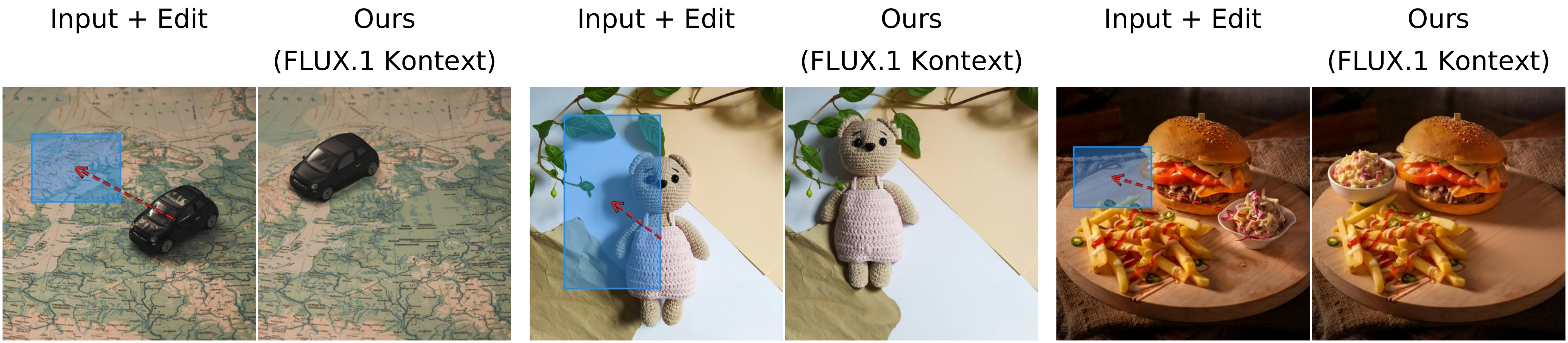}
    \caption{Our method generalizes to other DiT-based editing models such as FLUX.1 Kontext \cite{labs2025flux1kontextflowmatching}.}
    \label{fig:flux}
\end{figure}

\paragraph{Generalization to other DiT-based editing models.}
Our method is not tied to a specific backbone. Fig.~\ref{fig:flux} shows that the same RoPE-based spatial warping and depth conditioning generalize to FLUX.1 Kontext~\cite{labs2025flux1kontextflowmatching}, yielding comparable object relocation behavior.

\begin{figure}
    \centering
    \includegraphics[width=1.0\linewidth,
                     trim=0cm 16cm 0cm 0cm,
                     clip]{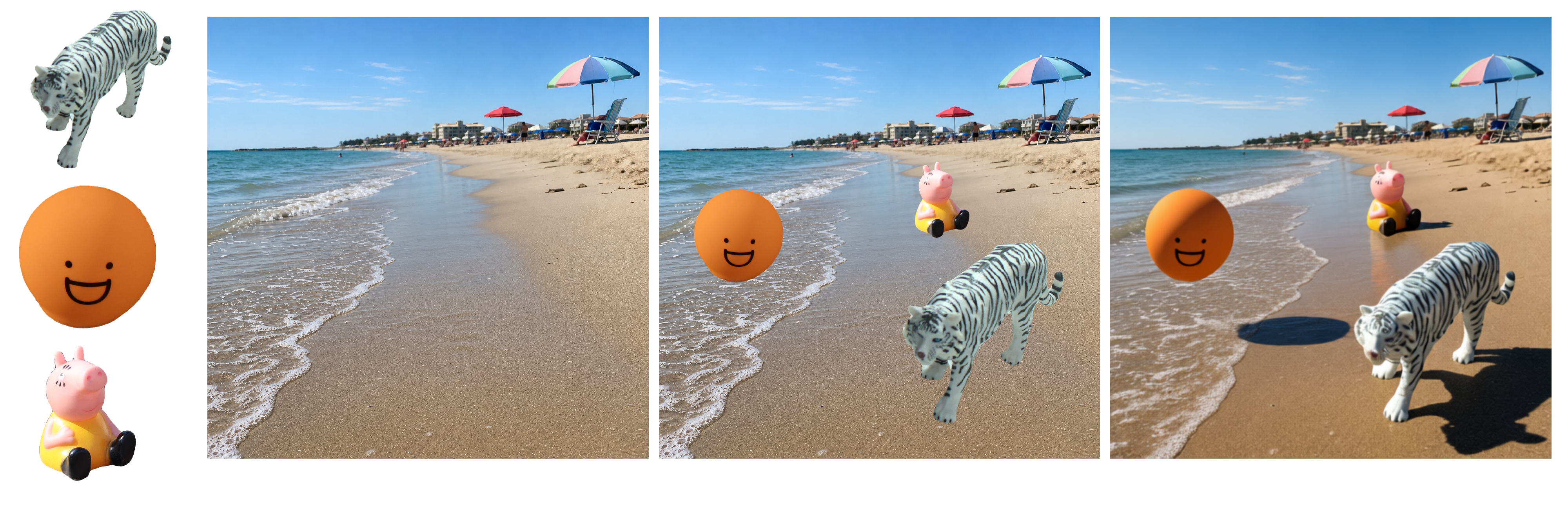}
    \caption{Given an image with desired objects copy-pasted on top, our method performs object addition.}
    \label{fig:image_enhancement}
\end{figure}

\paragraph{Object addition.}
Beyond relocation, our method also supports object addition. Given an input image with desired objects copy-pasted on top, the model realistically blends them into the scene, harmonizing lighting, shadows, and contact with surrounding surfaces (Fig.~\ref{fig:image_enhancement}).

\subsection{User Study}\label{subsec:user_study}

\begin{figure}[t]
    \centering
    \includegraphics[width=\linewidth]{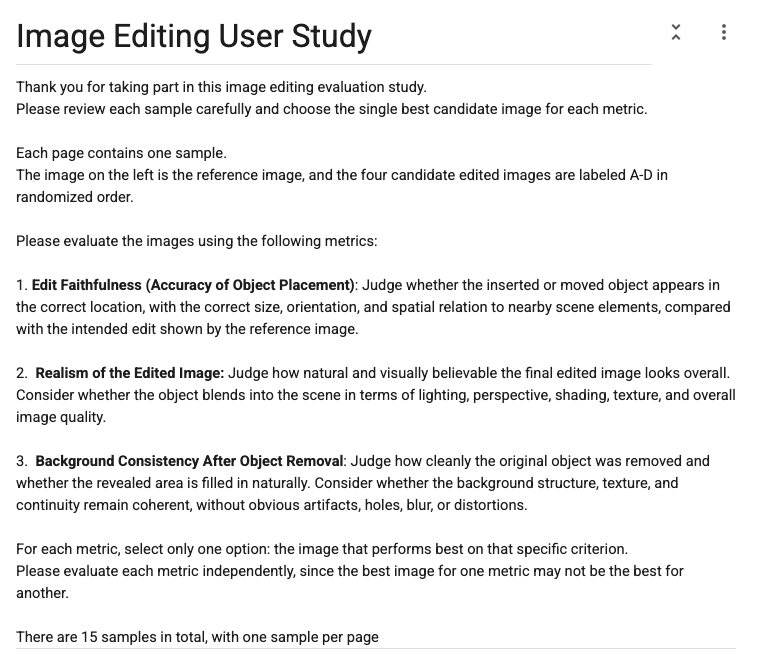}
    \caption{Instructions shown to participants for completing the user study.}
    \label{fig:user_study}
\end{figure}

We conducted a user study to compare our method against competing approaches. A total of 20 volunteers participated, and we evaluated 15 randomly selected samples without cherry-picking. To keep the study tractable, we restricted the comparison to the three strongest baselines: FreeFine~\cite{zhu2025training}, ObjectMover~\cite{yu2025objectmover}, and Qwen-Edit~\cite{wu2025qwenimagetechnicalreport}. For each sample, participants were asked to select the best result along three axes: edit faithfulness (accuracy of object placement), realism of the edited image, and background consistency after object removal. The order of methods was randomized for each question, and no time limit was imposed. The instructions shown to participants are provided in Figure~\ref{fig:user_study}, and the aggregated voting results are shown in Table~\ref{tab:user_study}. Our method is preferred by a clear majority of participants across all three metrics, receiving 83.4\% of votes overall.

\begin{table}[ht]
\centering
\caption{User study results. 20 respondents evaluated 15 samples across 3 metrics, yielding 300 votes per metric and 900 votes overall. Values show percentage of votes (vote count in parentheses). Best results in \textbf{bold}.}
\label{tab:user_study}
\setlength{\tabcolsep}{4pt}
\begin{tabular}{lcccc}
\toprule
Metric & Ours & ObjectMover & FreeFine & Qwen-Edit \\
\midrule
Edit Faithfulness       & \textbf{82.3\%} (247) & 7.3\% (22) & 10.3\% (31) & 0.0\% (0) \\
Realism                 & \textbf{82.7\%} (248) & 7.7\% (23) & 7.3\% (22)  & 2.3\% (7) \\
Background Consistency  & \textbf{85.3\%} (256) & 6.7\% (20) & 6.3\% (19)  & 1.7\% (5) \\
\midrule
Overall                 & \textbf{83.4\%} (751) & 7.2\% (65) & 8.0\% (72)  & 1.3\% (12) \\
\bottomrule
\end{tabular}
\end{table}

\clearpage

\end{document}